%% file: MA-IRL-SC.tex
\documentclass[journal]{IEEEtran}

\usepackage[utf8]{inputenc}
\usepackage{amssymb}
\usepackage{amsmath}
\usepackage{amsthm}
\usepackage{array}
\usepackage{bm}

\usepackage{xcolor}
\usepackage{graphicx}
\usepackage[colorlinks,citecolor=blue]{hyperref}
\usepackage{subcaption}
\usepackage{booktabs,etoolbox} 
\usepackage{stfloats} 

\usepackage{graphicx}
\usepackage{algorithm}
\usepackage{algpseudocode}
\usepackage[acronyms, shortcuts]{glossaries}
\usepackage{comment}
\usepackage[T1]{fontenc}

\let\oldac\ac
\renewcommand*\ac[1]{\textcolor{black}{\oldac{#1}}}

\usepackage{multirow}

\ifCLASSINFOpdf
\else
\fi

\theoremstyle{definition}

\theoremstyle{remark}

\def\keywords{\normalfont%
    \if@twocolumn%
    \@IEEEabskeysecsize\bfseries\textit{Index Terms}---\,\relax%
}

\begin{document}

\title{Pride and Prejudice: Toward an Information-Theoretic Framework for Mutually Communicative Driver Behavior Modeling
}

\author{Tingjun Li$^{1}$,
        Nan Xu$^{1}$, 
        Shuo Feng$^{2}$, 
        Hassan Askari$^{3}$,
        Bruno Henrique Groenner Barbosa$^{4}$
        and
        Konghui Guo$^{1}$
        \thanks{This work was supported in part by the National Key R\&D Program of China under Grant 2023YFB2504400; in part by the National Natural Science Foundation of China under Grant 52372385. (Corresponding author: Nan Xu)}
        \thanks{T. Li, N. Xu, and K. Guo are with the State Key Laboratory of Automotive Chassis Integration and Bionics, Jilin University, 5988 Renmin Street, Changchun, 130025, Jilin, China (e-mail: litj21@mails.jlu.edu.cn; xunan@jlu.edu.cn; guokh@jlu.edu.cn). } 
        \thanks{S. Feng is with the Department of Automation, Beijing National Research Center for Information Science and Technology, Tsinghua University, Beijing, China (e-mail: fshuo@tsinghua.edu.cn). }
        \thanks{H. Askari is with the Department of Engineering, Brock University, 1812 Sir Isaac Brock Way, St. Catharines, L2S3A1, Ontario, Canada (e-mail: haskari@brocku.ca). }
        \thanks{B. Barbosa is with the Department of Automatics, Federal University of Lavras, Av. Norte UFLA, Lavras, 37200-000, MG, Brazil (e-mail: brunohb@ufla.br). }}

\maketitle

\begin{abstract}

  Mixed autonomy driving becomes unsafe and inefficient when autonomous vehicles (AVs) and human-driven vehicles (HVs) misread each other's intentions. We study this problem as implicit mutual communication in lane changes. The proposed framework models how the ego vehicle both expresses its intent and probes the other driver's preference under epistemic uncertainty. It combines a level-k Bayesian persuasion game with virtual features for proactive signaling, information-theoretic rewards for mutual communication, and adaptive weights of communication affordances. We further introduce the Pride-Inquiry (P-I) and Pride-Prejudice (P-P) planes to analyze communication intensity and tendency.

  The model is calibrated with a Communication-Based Multi-Agent Inverse Reinforcement Learning algorithm (C-MIRL) on the naturalistic NGSIM dataset. Compared with the non-communicative baseline, the proposed model reduces the prediction error of mandatory lane changes by up to 20\% while maintaining strong generalization. Driver-In-the-Loop questionnaire scores are positively correlated with the calibrated communication variables, supporting the subjective validity of the model. The learned rewards further show that inquiry and listening affordances contribute more than pride and expression alone, and that inquiry preference varies more strongly across drivers. These results support explicit modeling of mutual communication and epistemic uncertainty in interactive driving.

  \end{abstract}



\IEEEpeerreviewmaketitle

\section{Introduction}\label{sec:intro}
\input{sections/1-intro.tex}

\section{Related Work}\label{sec:rel}
\input{sections/2-literature-review.tex}

\textcolor{black}{
\section{Methodology}
\input{sections/3-0-structs.tex}

}

\section{Problem Formulation}\label{sec:prblm}
\input{sections/3-methods.tex}

\section{Bayesian Persuasion}\label{sec:bp_ip}
\input{sections/3-1-game_struct.tex}

\input{sections/3-1-L_0.tex}
\input{sections/3-2-L_1_mod_featr.tex}
\input{sections/3-3-L_2_func_rewards.tex}
\input{sections/3-4-info_rewards.tex}



\section{Experimental Validation}
\label{sec: exp-vali}
\input{sections/4-0-mirl.tex}
\input{sections/5-1-data-2-lf.tex}

\section{Results and Discussions}\label{sec:exp-res}
\input{sections/5-exp-res.tex}

\section{Conclusion}\label{sec:concl}
\input{sections/6-discussion.tex}

\begin{appendices}
\section{Relationship between inquiry and clarity}
\input{appendix/1-inquiry-as-eu.tex}
\end{appendices}

\section*{Acknowledgement}
This work was supported in part by the National Key
Research and Development Program of China under Grant 2023YFB2504400
and in part by the National Natural Science Foundation of China under Grant 52372385. (Corresponding author: Nan Xu.)

\ifCLASSOPTIONcaptionsoff
  \newpage
\fi

\bibliography{MC-BIRL.bib}
\bibliographystyle{IEEEtran}





\begin{IEEEbiography}[{\includegraphics[width=1in,height=1.25in]{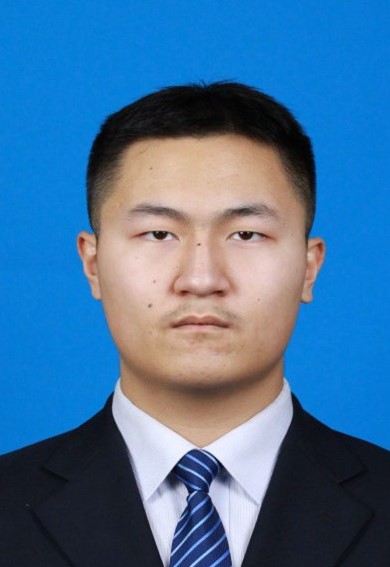}}]{Tingjun Li}
  received the B.E. degree in automobile service engineering from Jilin University, Hohhot, China, in 2019. He is currently pursuing the Ph.D. degree in automotive engineering from Jilin University, Changchun, China. His current research focuses on interactive decision-making for socially compatible and trustworthy autonomous driving.
 \end{IEEEbiography}
 \vspace{0.2cm} 
\begin{IEEEbiography}[{\includegraphics[width=1in,height=1.25in]{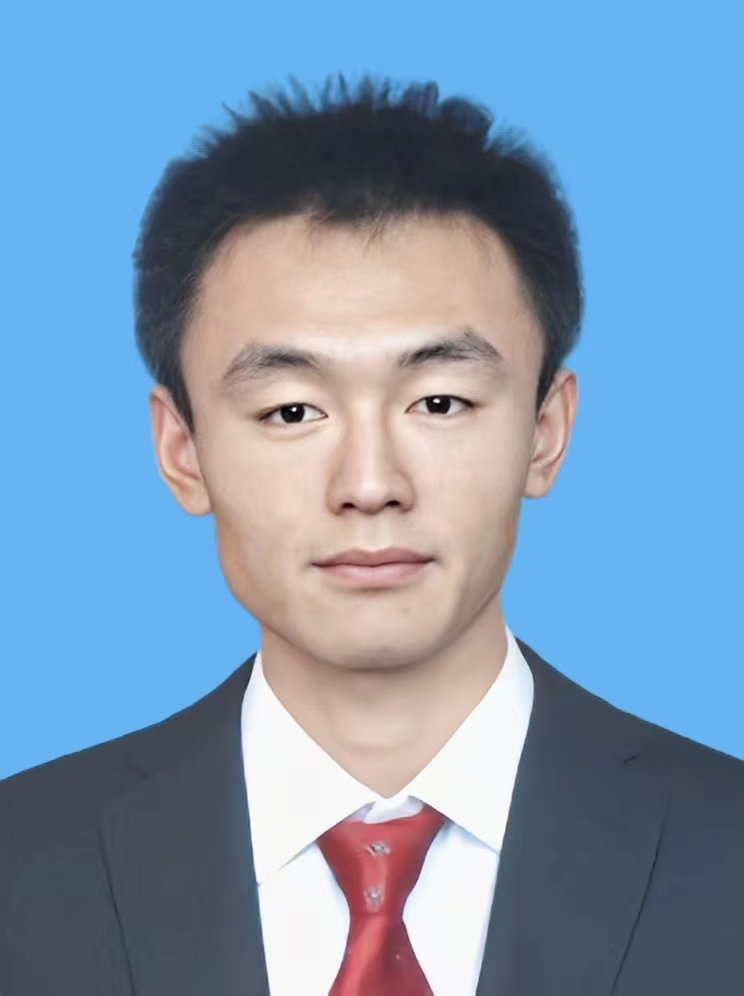}}]{Nan Xu}
 received the Ph.D degree in 2012 from Jilin University, Changchun, China. He is currently a Professor with the State Key Laboratory of Automotive Simulation and Control, Jilin University of China, and became the Tang Aoqing Professor in Jilin University from 2020. In 2019, he was a Visiting Scholar with the Department of Mechanical and Mechatronics Engineering, University of Waterloo. His current research focuses on tire dynamics and control, intelligent tire, vehicle state estimation, and dynamics and stability control of electric vehicles and autonomous vehicles.
\end{IEEEbiography}
\vspace{0.2cm} 
\begin{IEEEbiography}[{\includegraphics[width=1in,height=1.25in]{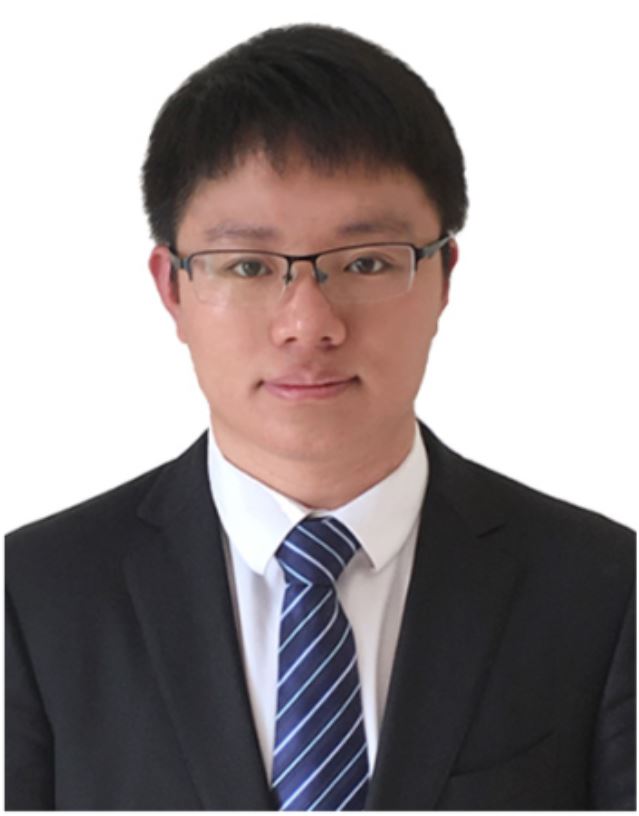}}]{Shuo Feng}
 received the bachelor's and Ph.D. degrees from the Department of Automation, Tsinghua University, China, in 2014 and 2019, respectively. He was also a joint Ph.D. student in civil and environmental engineering at the University of Michigan, Ann Arbor, MI, USA, from 2017 to 2019, where he was a Post-Doctoral Researcher from 2019 to 2021 and an Assistant Research Scientist from 2021 to 2023. 
 He is currently an Associate Professor with the Department of Automation, Tsinghua University, China. 
 His current research interests include connected and automated vehicle evaluation, mixed traffic control, and transportation data analysis.
\end{IEEEbiography}
\vspace{0.15cm}  
\begin{IEEEbiography}[{\includegraphics[width=1in,height=1.25in]{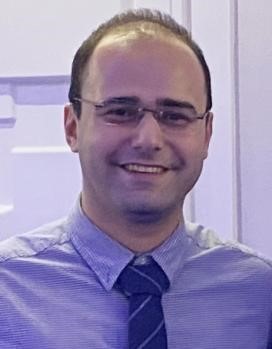}}]{Hassan Askari}
was born in Rasht, Iran. He received
the Ph.D. degree in mechanical and mechatronics engineering from the University of Waterloo,
Waterloo, ON, Canada, in 2019.
He has coauthored one book and one book chapter
published by Springer. He has authored or coauthored more than 70 journal and conference papers
in the areas of nonlinear vibrations, applied mathematics, nanogenerators, and self-powered sensors.
Dr. Askari is the recipient of several prestigious
awards, including the Outstanding Researcher at the
Iran University of Science and Technology, the Fellowship of the Waterloo
Institute of Nanotechnology, Waterloo, the Ontario Graduate Scholarship, and
the University of Waterloo President Award. He is an active Reviewer of more
than 40 journals and is an Editorial Board Member of several scientific and
international journals.
\end{IEEEbiography}
\vspace{0.15cm}  
\begin{IEEEbiography}[{\includegraphics[width=1in,height=1.25in]{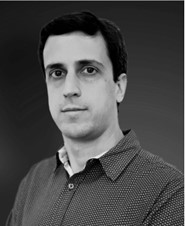}}]{Bruno Henrique Groenner Barbosa }
received the B.Sc. degree in control and automation engineering,
and the M.Sc. and Ph.D. degrees in electrical engineering from the Federal University of Minas Gerais,
Belo Horizonte, Brazil, in 2003, 2006, and 2009,
respectively.
In 2008, he was a Visiting Researcher with the
University of New South Wales, Sydney, NSW,
Australia. In 2020, he was a Visiting Professor with
the Department of Mechanical and Mechatronics
Engineering, University of Waterloo, Waterloo, ON,
Canada. He is an Associate Professor with the Department of Automatics,
Federal University of Lavras, Lavras, Brazil, where he heads the Artificial
Intelligence and Automation Research Group. His research interests include
artificial intelligence techniques, such as neural networks and evolutionary
algorithms, for solving real-world engineering problems. These have included
identification of nonlinear dynamical systems, development of soft sensors,
pattern recognition, and system optimization and control.
\end{IEEEbiography}
\vspace{0.05cm}  
\begin{IEEEbiography}[{\includegraphics[width=1in,height=1.25in]{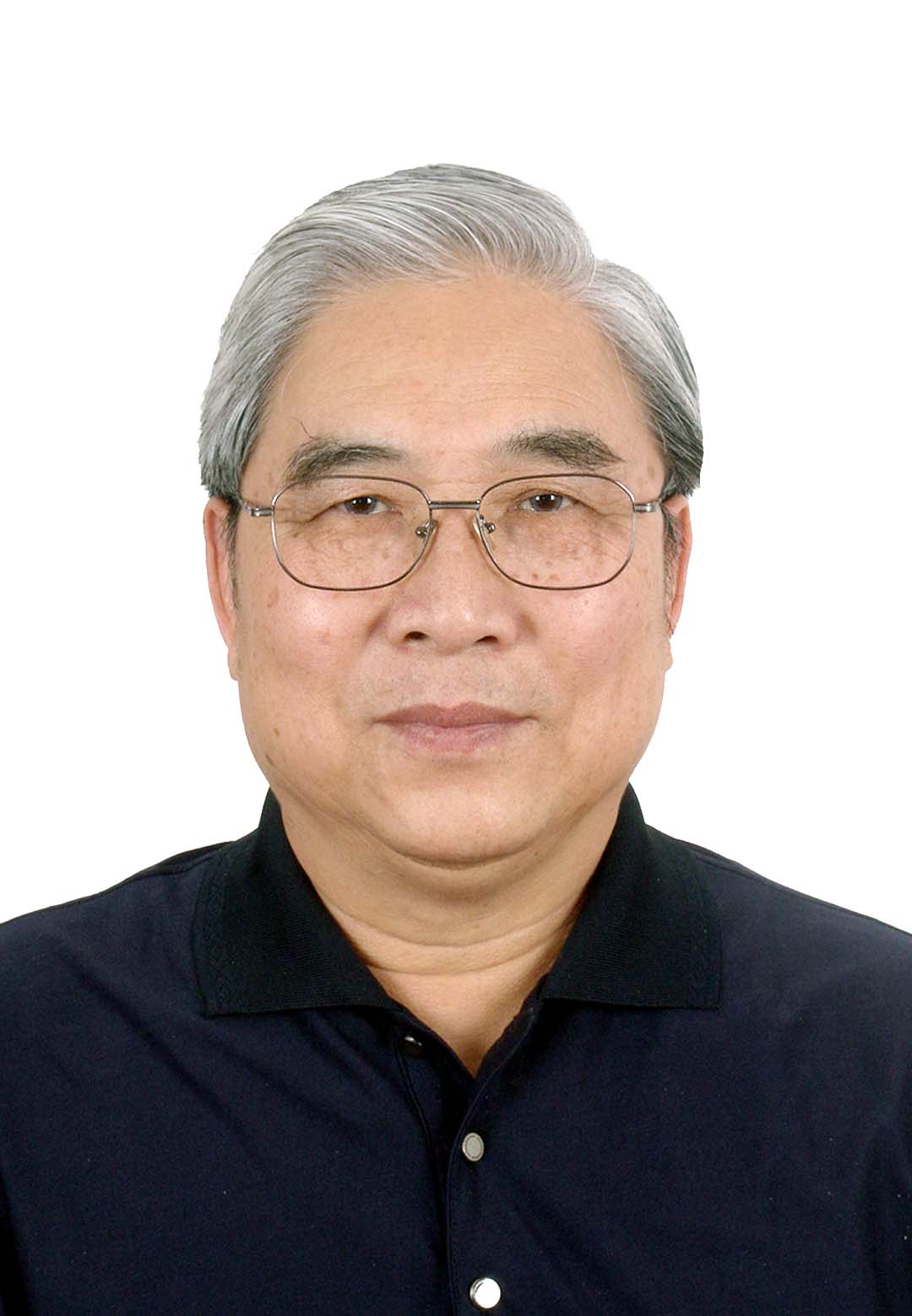}}]{Guo Konghui}
  received the B.Sc. degree in automobile and tractor from Jilin University of Technology in 1956. He was named a doctoral supervisor in 1990 and was elected as an academician of the Chinese Academy of Engineering in 1994. He is currently the honorary dean of the School of Automotive Engineering of Jilin University, a member of the editorial board of IATSS Research and Vehicle System Dynamics. He was the chief engineer of Changchun Automobile Research Institute, a visiting researcher at the University of Michigan, the chairman of the technical committee of the 5th Annual Conference of the International Society of Automotive Engineers of the Pacific, and the chairman of the technical committee of the 25th Annual Conference of the World Society of Automotive Engineers (FISITA).
\end{IEEEbiography}


\end{document}

%% file: sections/1-intro.tex
\textcolor{black}{It is a truth universally acknowledged that automation bias in human-machine systems necessitates clear communication between humans and machines to maintain effective control \cite{kahn2024-aisafetyautomationbias}.} Therefore, clear communication is essential when autonomous vehicles (AVs) interact with human-driven vehicles (HVs) in safety-critical traffic scenarios. Autonomous driving can improve safety and road efficiency \cite{xu2021-calibrationevaluationresponsibilitysensitive}, but more than 31\% of AV disengagements are still linked to failures in anticipating other road users or to unwanted maneuvers \cite{departmentofmotorvehicles2024-autonomousvehicledisengagement}. These failures are not only a perception problem; they also arise when human drivers and AVs misread each other's intentions.

This issue is especially visible in mixed traffic. AVs may behave in ways that humans find unusual or hard to interpret. Misunderstanding also wastes right-of-way and reduces traffic efficiency. At the same time, human drivers may exploit AV conservatism and bully them in interaction \cite{sun2024-greaterprosocialityother}. These phenomena reflect not only social preference, but also epistemic uncertainty (EU) and biased beliefs about the other agent. In practice, drivers reduce this uncertainty through implicit communication, such as small changes in speed or trajectory that reveal intent \cite{lee2021-roadusersrarelya}.

EU describes ignorance about the model, whereas aleatoric uncertainty (AU) captures noise inherent in the data \cite{kendall2017-whatuncertaintieswe}. In driving, EU arises from limited training data \cite{shao2024-uncertaintyawarepredictionapplication}, unknown environments such as occlusion \cite{sun2019-behaviorplanningautonomous}, and unknown driver traits such as aggressiveness \cite{zhang2022-humanlikeinteractivebehavior}. 
At the motion-control layer, robust tracking controllers have been developed to improve vehicle path-following performance under parametric uncertainty, such as robust curvature-preview drift control with STA-ISMC~\cite{gan2026-robustcurvaturepreview} and fuzzy adaptive dynamic MPC for lateral trajectory tracking~\cite{teng2026-robustlateraltrajectory}. 
On the other hand, planning under EU remains difficult \cite{hoel2023-ensemblequantilenetworks, yang2023-robustdecisionmakingautonomous, tang2022-predictionuncertaintyawaredecisionmakingautonomousb}. Existing approaches usually model EU with predictive distributions \cite{yang2023-predictionfailureriskawarea} or risk-sensitive objectives such as confidence \cite{zhou2023-dynamicallyconservativeselfdriving}. Here, we instead study how interaction itself can reduce EU through communication. 

This perspective motivates a mutually communicative view of AV-HV interaction. 
\textcolor{black}{Our research is founded on a structural assumption of mutual communication: agents communicate because they assume others are listening, and they listen because they assume others are communicating. This bidirectional assumption is grounded in embodied cognition theory, where communication occurs naturally through physical actions besides explicit signal exchange \cite{obayashi2025-embodiedintelligenceparadigm}. Our previous work has also established its correlation with human driver's trust through subjective evaluations \cite{li2025-modellingtrustinteraction}. }

For simplicity, only implicit communication is considered here. It occurs when a driver's actions affect motion while also conveying a message or request to other road users \cite{markkula2020-defininginteractionsconceptual}.

Existing AV interaction models range from reactive models based on current observations \cite{wan2014-modelingvehicleinteractionsa}, to affordance-based models that evaluate future desirability and viability \cite{pezzulo2016-navigatingaffordancelandscape, markkula2023-explaininghumaninteractions}, to communicative models that require proactive signaling \cite{siebinga2023-modellingcommunicationenabledtraffic}. In reactive models, the interaction is considered fully observable and EU is not considered. With affordance-based models, priority assertion is significantly less pronounced than what has been observed in human drivers \cite{markkula2023-explaininghumaninteractions}. Related work also combines game-theoretic planning with belief-space planning to exploit belief dynamics \cite{schwarting2021-stochasticdynamicgames}. However, context-based quantification of communication information in vehicle interaction remains limited. A mutually communicative model should distinguish persuasive tactics from self-interest preferences and explicitly address prejudice in mixed-autonomy interaction.

In this paper, we propose an information-theoretic framework for implicit mutual communication in lane-change interactions, a representative driving scenario with strong uncertainty and negotiation~\cite{yao2025-hierarchicalpredictionuncertaintyaware}. A level-k game with virtual features is used for Bayesian persuasion to model proactive signaling by the AV. \textcolor{black}{We further introduce a unified information-theoretic framework to model interaction uncertainties. This work extends our prior publication by modeling epistemic uncertainty (EU) in the information channel of belief and action, $p(a|\theta)$, complementing the previous focus on aleatoric uncertainty (AU) in the channel of action and reward, $p(r|a)$. } Communication rewards for pride, prejudice, and inquiry quantify information exchange and define two communication planes that describe engagement tendency and intensity. Finally, a communication-based multi-agent inverse reinforcement learning (C-MIRL) framework is adapted from our previous work~\cite{11048679} and validated on naturalistic driving data and Driver-In-the-Loop (DIL) experiments.

The contributions of the proposed model are threefold. 

\begin{enumerate}

    \item We establish a dynamic Bayesian persuasion framework based on level-k reasoning and virtual features. Within this framework, pride and prejudice are quantified by information gain, and the ego makes personalized expressive decisions. 
    
    \item We introduce a mutual-information inquiry model to quantify expected prejudice corrected under epistemic uncertainty. We also model communication affordances as adaptive preferences shaped by environmental conditions, and validate the framework with naturalistic driving data and DIL experiments. 
    
    \item We summarize communication behavior with two interpretable planes: the Pride-Inquiry (P-I) plane and the Pride-Prejudice (P-P) plane, which describe communication tendency and intensity.

\end{enumerate}

The remainder of this paper is organized as follows. Section II reviews related work. Sections III and IV present the problem formulation and Bayesian persuasion framework. Sections V and VI describe the calibration framework and experiments. Section VII reports the results, and Section VIII concludes the paper.

%% file: sections/2-literature-review.tex
For the most part, the literature review contains three themes: 
(a) algorithmic discrimination, (b) game-theory-based vehicular communicative protocol, and (c) active social learning for vehicle interaction. 

\subsection{Algorithmic Discrimination} 
\label{sub:algorithmic discrimination}

Discrimination can be introduced to algorithms during dataset construction and training. Biased or underrepresented data may provide poor foundations for algorithm training. In the training process, feature selection may introduce designer bias~\cite{chen2023-ethicsdiscriminationartificial}. Besides data augmentation, conventional approaches to mitigating representation bias rely on either resampling or reweighting data points. In more skewed datasets, domain adaptation and self-supervised learning are adopted~\cite{hu2024-inclusivedecisionmaking}. New developments in the machine learning literature suggest the use of adaptive learning to dynamically adjust the weight of each data point during model training~\cite{lin2020-focallossdense}. These adaptive machine learning methods offer the advantage of recalibrating resampling or reweighting at each training step. 
Specifically, to estimate object preference in vehicle interaction, preference dynamics are adopted in the adaptive learning process~\cite{schwarting2019-socialbehaviorautonomous}. This process can reweight data but may also reinforce stereotyping. Here, we adopt it to model pride and prejudice in vehicular implicit communication. 

For the evaluation of discrimination, prejudice and volatility are modeled with the most significant stereotype under different circumstances ~\cite{liu2024-prejudicevolatilitystatistical}. However, to focus on the most prejudiced candidate may lead to missing the bigger picture. 
Stereotyping behavior is driven by more values and preferences than self-interest ~\cite{becker1971economics}. We propose an information-theoretic model of stereotyping risk to measure the prejudice over the whole distribution. Further, it is modeled in a communication framework where its dynamics can be captured to fight epistemic uncertainty between interacting partners. 


\subsection{Game-Theory Based Vehicular Communicative Protocol} 
\label{sub:game theory based vehicular communicative protocol}


Game-theory-based communication frameworks used in vehicular implicit communication mainly encompass established models such as bargaining games \cite{nash1950bargaining}, signaling games \cite{spence1973-jobmarketsignaling}, and Bayesian persuasion \cite{kamenica2011-bayesianpersuasion}. Bargaining games have been adopted for lane-change interactions \cite{dossantos2019-bargaininggameapproach} and ramp metering \cite{heshami2021-rampmeteringcontrol}. However, the requirement for state and control information exchange is overly restrictive for non-connected environments.  
\cite{shao2020-discretionarylanechangingdecisionmaking} propose a signaling-game-based model with systematic utility design and three types of equilibria. However, the signaling process and sender type are assumed to be separable rather than quantified, which could be improved for implicit communication. 
Compared to these alternative communication models, Bayesian persuasion grants the sender greater authority to commit to their messaging strategy \cite{kamenica2019-bayesianpersuasioninformation}. However, to our knowledge, a comprehensive model of personalized expressive behavior that distinguishes between persuasive tactics and self-interest preferences has not been proposed to account for pronounced human behavior. We also propose pride exudation as the target of persuasion, which is to affect the second-order belief of objects.  

\subsection{Active Social Learning for Vehicle Interaction} 
\label{sub:active learning for vehicle interaction}

Proactive methods have been adopted in vehicle social learning to gather information more efficiently. 
In ~\cite{deng2022-lanechangedecisionmaking}, driver aggressiveness is estimated and integrated into the Bayesian game framework for lane change. Furthermore, data regarding the aggressiveness of nearby vehicles is systematically gathered by employing a tentative behavioral approach. ~\cite{zhang2022-humanlikeinteractivebehavior} proposes a field-based aggressiveness model and a probing behavior algorithm based on the gradient of the risk field. However, the effect of probing is not quantified. 

Active inverse learning approaches initially elicit informative human actions, record them in the dataset, and subsequently deduce the human's objective function, iterating this cycle as needed. These techniques ensure the informativeness of human actions by either maximizing the reduction in the volume of the hypothesis space \cite{basu2019active} or optimizing information gain \cite{cui2018-activerewardlearning, daniel2015-activerewardlearning}. \cite{yu2023-activeinverselearning} further considers non-cooperative interactions and infers information from actions by adopting the KL divergence of state distributions under different hypotheses and maximizing the worst-case pairwise distance. However, pairwise distance captures only part of the information gain. \cite{engstrom2024-resolvinguncertaintyfly} employs expected information gain in the expected free energy of active inference to quantify epistemic uncertainty under occlusion. Similarly, we maximize information gain from object action to preference in implicit communication during lane changes. In this paper, we introduce and validate an information-theoretic framework of implicit mutual communication, combining insights from ego actions and object beliefs within the Bayesian persuasion framework.

%% file: sections/3-0-structs.tex
\label{sub:Method}

\begin{figure}[ht]
    \centering
    \includegraphics[width=\linewidth, trim=90 50 80 82, clip]{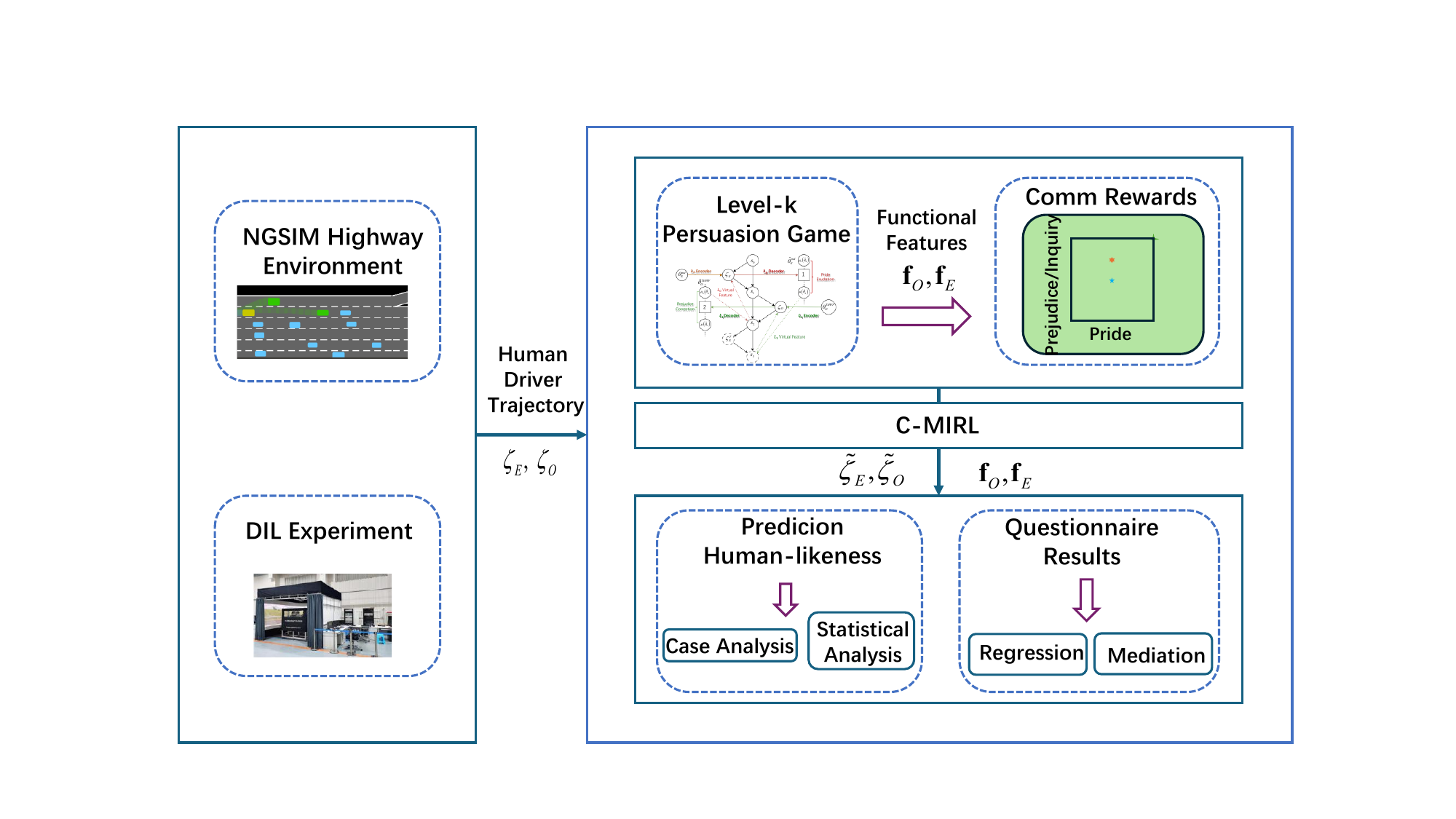}
    \caption{Detailed illustration of the communication preference calibration framework. Complete flow chart of the proposed communication-based calibration framework. The pipeline integrates: (1) human demonstration trajectories from NGSIM/DIL experiments as input; 
    (2) level-k persuasion game structure; (3) communication utilities and communication planes; (4) C-MIRL optimization to iteratively calibrate functional weights; 
    (5) objective and subjective evaluation. 
    }
    \label{fig:pp_mdl}
\end{figure}

To establish a communication-aware decision framework under cognitive uncertainty, we propose a level-k persuasion game with communication rewards and embed it into the MIRL framework. Fig.~\ref*{fig:pp_mdl} presents the overall pipeline of the proposed C-MIRL framework. 
We first formulate the decision-making problem as a Bayesian persuasion game problem. 
Then a communication model is constructed with level-k games. 
Further, functional rewards and communication preferences are used to capture the human belief-updating mechanism. 
Finally, we embed this mechanism into hierarchical multi-agent IRL and validate the calibrated policy through objective trajectory prediction and subjective human evaluation.

%% file: sections/3-methods.tex
We study a two-vehicle lane-change interaction and model both the ego vehicle (denoted by $()_E$) and its belief about the object vehicle in the target lane (denoted by $()_O$). The ego observes the interaction state $\mathbf{x}_t \in \mathcal{X}$. 
The observation is used to predict the probabilities of trajectories $\zeta \in \mathcal{Z} $
of the interacting vehicles with the assumption of a discrete-time setup and a finite time horizon $T$.

A state lattice planner \cite{gautam2024-overviewmotionplanningalgorithms} is adopted for each vehicle. \textcolor{black}{The two-vehicle lane-change interaction setup and trajectory sampling framework follow our previous work \cite{11048679}. } Both vehicles are modeled as noisy optimizers that seek to maximize their cumulative rewards $R_E$ and $R_O$ over the sampled trajectories. 

To model nonverbal mutual communication, we adopt Bayesian persuasion. 

Bayesian persuasion involves stronger commitment than cheap-talk games \cite{crawford1982strategic} or signaling games \cite{spence1978job}, which have been used in discretionary lane-changing models \cite{shao2020-discretionarylanechangingdecisionmaking}. In this framework, the ego (sender) commits to a message distribution $s \in S$ conditioned on the world state $\omega \in \Omega$ so that it can influence the object's (receiver's) belief $\mu$ and action $a$ \cite{kamenica2019-bayesianpersuasioninformation}. This mapping is the signal $\pi$. 

Given a signal realization $s$, a posteriori belief \textcolor{black}{ $\mu_s(\omega)$} is induced  
\begin{equation}
    \mu_s(\omega)=\frac{\pi(s \mid \omega) \mu_0(\omega)}{\sum_{\omega^{\prime} \in \Omega} \pi\left(s \mid \omega^{\prime}\right) \mu_0\left(\omega^{\prime}\right)}
\end{equation}

\textcolor{black}{where the sender and receiver are assumed to share a prior $\mu_0 \in \text{int} (\Delta (\Omega))$.}
Accordingly, each signal leads to a distribution over posterior beliefs, which is denoted by $\tau$

\begin{equation}
    \tau(\mu)=\sum_{s: \mu_{\pi(\mid s)}=\mu} \sum_{\omega^{\prime} \in \Omega} \pi\left(s \mid \omega^{\prime}\right) \mu_0\left(\omega^{\prime}\right)
\end{equation}

The sender optimizes its continuous utility $v(a,\omega)$ subject to the receiver utility $u(a,\omega)$. 

Therefore, the ego needs to solve

\begin{equation}
\begin{gathered}
    \max _\tau \mathbb{E}_{\mu \sim \tau} \hat{v}(\mu)\\
    \text{ subject to } \mathbb{E}_{\mu \sim \tau} \mu=\mu_0
\end{gathered}
\end{equation}

Any posterior distribution induced by a signal satisfies this constraint. This property is called Bayesian plausibility and shows how the prior constrains the induced beliefs. 

We then build a level-k interaction model and introduce information rewards to capture the ego's expression and listening preferences.

%% file: sections/3-1-game_struct.tex
\subsection{Bayesian Persuasion with Level-2 Game}\label{sec:3-1-bp_ip}

A level-k game is used to model mutual communication in the interaction. Agents are assumed to reason at different cognitive levels $k \in N$. A level-0 agent follows a fixed policy without strategic interaction. A level-1 agent best-responds to that behavior, and a level-$k$ agent best-responds under the assumption that the others are level-$(k-1)$ agents \cite{li2018-gametheoreticmodeling}.

\begin{figure}
    \centering
    \includegraphics[scale=0.35, trim=120 0 120 0, clip]{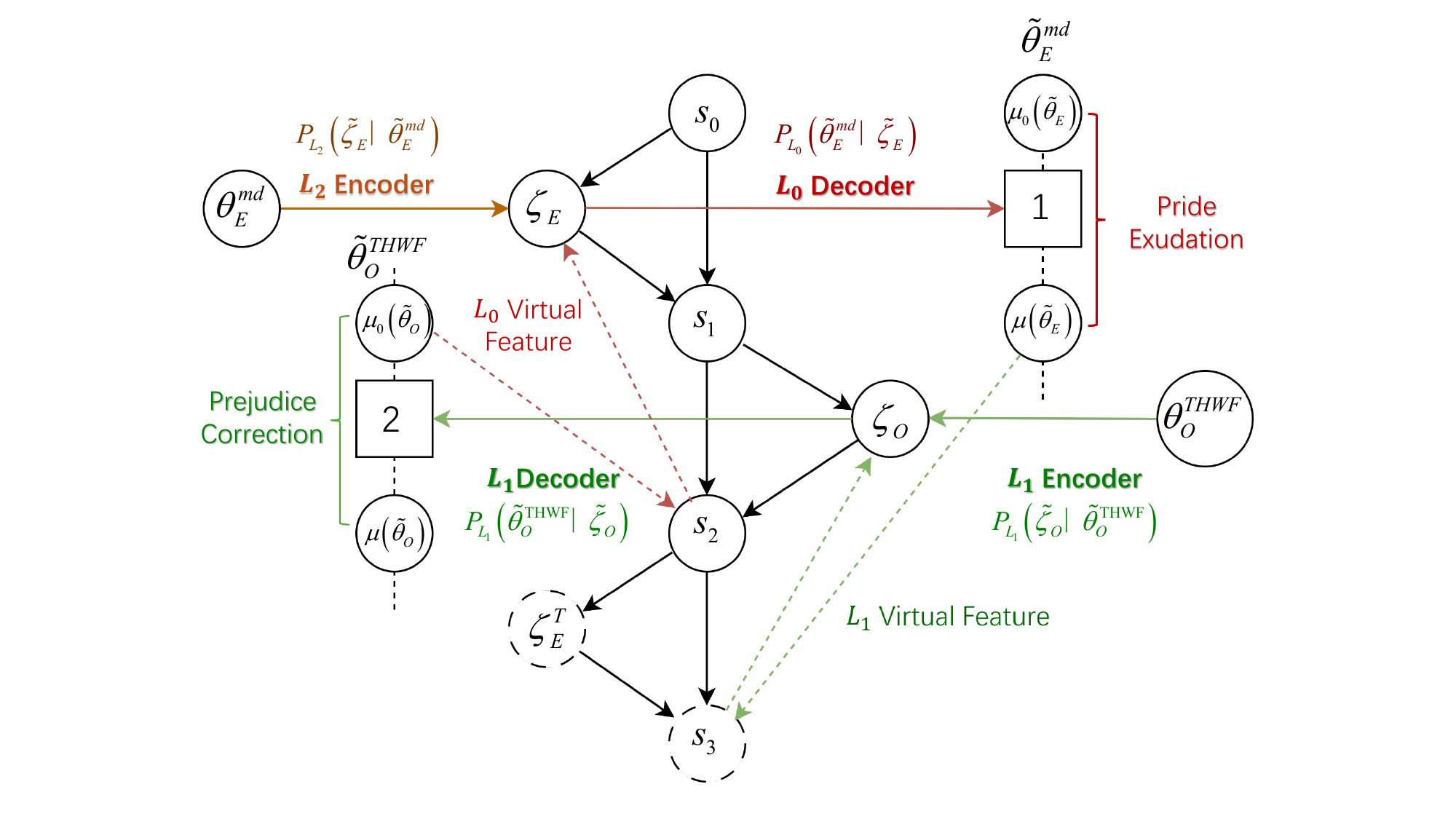}
        \caption[short]{The Bayesian persuasion with level-2 2-step game structure and the bidirectional information channels enabling mutual ego-object communication. 
        \textcolor{black}{Solid circles represent actual states and actions taken by agents and sampled in the simulator; dashed circles represent hypothetical states considered during reasoning.         
        Solid arrows show causal dependencies (action transitions) and information flow (belief updates), whereas dashed arrows illustrate how belief shapes these dynamics by influencing the prospect through virtual features.
        }}
        \label{fig:1}
    \end{figure}

We model Bayesian persuasion as a level-2, two-step dynamic game, shown in Fig. \ref*{fig:1}. Nature first selects the interacting vehicles' preferences. The ego then sends a signal through its trajectory choice $\zeta_E$, which expresses $\tilde{\theta}^{md}_E$, the second-order belief over mandatory reward preference. The object responds by choosing $\zeta_O$ after inferring the distribution of $\tilde{\theta}^{md}_E$, expressing $\tilde{\theta}^\text{THWF}_O$, and predicting the ego's next move $\zeta^T_E$ over horizon $T$. Here, $\tilde{\theta}^\text{THWF}_O$ denotes the first-order belief over the frontal time-headway preference of the level-1 object.

In this process, the ego serves as the Sender and optimizes its reward as a level-2 agent, with an $L_2$ encoder. 

\begin{equation}
\zeta_E^{*}=\arg \max _{\zeta_E} \mathbb{E}_{\tilde{\theta}^{md}_E \sim \mu, \tilde{\theta}^\text{THWF}_O \sim \mu_0}R_E\left(\mathbf{x}^0, \zeta_E,\tilde{\theta}^{md}_E, \tilde{\theta}^\text{THWF}_O \right)\\
\end{equation}

where $\mu_0$ is the prior belief distribution and $\mu$ is the posterior. By choosing $\zeta_E$, the ego influences both the object's updated belief $\mu(\tilde{\theta}^{md}_E)$ and the resulting object action $\zeta_O$, and therefore its own reward. 

The receiver computes the posterior belief with Bayes' theorem using its level-0 decoder model of the ego. 

\begin{equation}
\begin{aligned}
\mu \left(\tilde{\theta}^{md}_E\right) 
= \frac{P_0(\tilde{\theta}^{md}_E)P_{L_0}\left(\tilde{\zeta}_E\mid \tilde{\theta}^{md}_E\right)}{P_{L_0}\left(\tilde{\zeta}_E\right)}
\end{aligned}
\end{equation}

\textcolor{black}{where $P_{L_0}\left(\tilde{\zeta}_E\right)$ denotes the probability of the ego model of level-0 choosing trajectory $\tilde{\zeta}_E$. }

The expression process and the listening process together form mutual communication. We use pride and prejudice to describe the corresponding information gain during this exchange. Virtual features are introduced in the $L_0$ and $L_1$ encoders to capture how current signaling changes future uncertainty and therefore meaning. 

\textcolor{black}{
It is important to clarify that our framework does not assume all drivers are explicitly willing to communicate. Rather, we model the reality that in interactive driving scenarios, all actions inevitably convey information, regardless of intent. Even a driver who is "unwilling to communicate" still signals their preferences through their speed, acceleration, and position choices. 
The Bayesian persuasion framework models how the receiving agent interprets these unavoidable signals, irrespective of the sender's conscious communication intent. This perspective is supported by empirical evidence from naturalistic driving data, which shows substantial heterogeneity in communication among drivers \cite{markkula2020-defininginteractionsconceptual}.}

%% file: sections/3-1-L_0.tex
\subsection{Level-0 Game with Virtual Features}\label{sec:3-1-L0}

In a rolling horizon framework, the level-0 object is perceived as a follower seeking to maximize the function:

\begin{equation}
    \zeta_E=\arg \max _{\zeta_E} R_E\left(\mathbf{x}^0, \zeta_E, \zeta_O^0\right)
\end{equation}

Here, $\mathbf{x}^0$ represents the initial state, $\zeta_E$ denotes a specified trajectory of the ego, and $\zeta_O^0$ signifies the given object trajectory at constant speed. According to the Boltzmann noisily-rational model, the probability of the interacting ego vehicle choosing its trajectory is exponential to the cumulative reward of the trajectory \cite{luce1959-individualchoicebehavior}.

\begin{equation}\label{eq:3}
    \zeta_E\left(\mathbf{x}^0, \zeta_O^0\right) \sim 
    P_{L_0}\left(\zeta_E \mid \mathbf{x}^0, \zeta_O^0\right) \propto e^{\beta R_e\left(\mathbf{x}^0, \zeta_O^0\right)}
\end{equation}

\textcolor{black}{ where $\beta$ regulates the rationality of the driver in the Boltzmann model, with higher values indicating more deterministic behavior. Here, we assume $\beta=1$ to form a softmax function following the practice in maximum entropy inverse reinforcement learning to ensure model identifiability \cite{ziebart2008-maximumentropyinverse}. }

\begin{equation}\label{eq:probs}
    P_{L_0}\left(\zeta_E \mid \mathbf{x}^0, \zeta_O^0\right)
    =\frac{e^{R_E\left(\mathbf{x}^0, \zeta_E, \zeta_O^0 \right)}}{\sum_{\tilde{\zeta}^i \in \mathbf{D}_E} e^{R_E\left(\mathbf{x}^0, \tilde{\zeta}_E, \tilde{\zeta}^i\right)}}
    \end{equation}
    
where \(R_E\left(\mathbf{x}^0, \zeta_E, \zeta_O^0 \right)\) denotes the ego vehicle's functional rewards. 
$R_E$ in the horizon of \(t\) is defined with the current state \(\mathbf{x}^0\), the candidate trajectory pair \(( \zeta_E, \zeta_O)\)

\begin{equation} \label{eq:6}
R_{E}\left(\mathbf{x}^0, \zeta_E, \zeta_O^0\right) = \boldsymbol{\theta_{E}}^T \sum_{t=0}^{N-1} \mathbf{f}_{E}\left(\mathbf{x}^t, \zeta_E^t, \zeta_O^t\right) 
\end{equation}

where \(\mathbf{f}_{E}  \in {R^9}\) denotes nine lower-level functional rewards. 
\(\boldsymbol{\theta_E } \in {R^9}\) denotes their weights. 


\begin{table}
    \centering
    \begin{tabular}{lllllllll}
        \toprule
        Symbol           & Meaning & Unit   \\
        \midrule
        v            &  velocity    &   $\text{m/s}$         \\
        $a_x$        &    longitudinal acceleration      &      $\text{m/s}^2$   \\
        $a_y$        &    lateral acceleration      &      $\text{m/s}^2$   \\
        $jerk$        &    longitudinal jerk      &      $\text{m/s}^3$   \\
        THWF        &    time headway to preceding vehicle      &     s  \\
        THWB        &    time headway to following vehicle      &    s\\
        collision    &   whether there is collision      &     \/  \\
        SI    &   Social Impact      &     $\text{m/s}^2$  \\
        \bottomrule
    \end{tabular}
    \caption{The functional features of interacting vehicles. }
\label{table:func_ft}
    \end{table}
    
Here, besides mandatory lane change preference, we also implement functional features that prioritize efficiency
, comfort, and energy conservation
, as well as safety, as summarized in Table \ref*{table:func_ft}
\cite{huang2022-drivingbehaviormodeling}.

Specifically, the exponentials of the negative time headway to the preceding vehicle THWF and following vehicle THWB are adopted as risk aversion features, assuming constant speed. 

In addition, there exists a penalty 
that is described as the expected deceleration of surrounding vehicles due to the behavior of the ego vehicle \cite{huang2022-drivingbehaviormodeling}. 

Furthermore, we propose a virtual feature model to describe the influence of belief over the rewards of the object. 
The key idea is to map the ego to its target lane and convert the lane-changing problem to a virtual car-following problem. 
Through this lens, a mandatory lane-changing utility considering future safe factors is designed to describe its urgency. 

\begin{equation}
    \label{eq17}
    f_{md}(\mathbf{s}_t) = 
    e^{\mathbb{E}_{\tilde{\theta}_O^{\text{THWF}}}C_{SF}} ({L_{C_{tmn}}-x_E})/{L_{T_{tmn}}},
    \end{equation}
where $L_{T_{tmn}}$ and $L_{C_{tmn}}$ denote the terminal positions of the target and current lanes, respectively, and $x_E$ denotes the current position. The closer the ego is to the merging area, the more advantageous it becomes to switch to the target lane. $C_{SF}$ denotes the safety factor at the end of the trajectory $\tilde{\zeta}^j$, which will be elaborated in \ref{sf}. The idea is that a lane change is more likely to occur and is thus preferred when better conditions are present. For simplicity, discretionary lane changes are calibrated with a model similar to that used for mandatory lane changes. Their presumed mandatory functional reward and related information expression reward are validated with experimental results.

The trajectory features are standardized through normalization to maintain coherence and reliability.
Moreover, the collision feature is consistently weighted at -10, with a maximum weight of -0.01 allocated to the time headway to the preceding vehicle and the social impact of lane-changing. This deliberate imposition of a negative weight is intended to enhance the safety of the model while decreasing causal confusion. 

%% file: sections/3-2-L_1_mod_featr.tex
\subsection{Level-1 Game with Virtual Features}\label{sec:3-2-bp_ip}

In Bayesian persuasion, the rewards of the object should be a function of the induced posterior belief distribution $\mu \left(\tilde{\theta}^{md}_E\right)$.
However, simulated features of the object used to compute rewards $R_O$ are certain given the trajectory of the ego. 

\begin{equation}
    \begin{gathered}
    P_{L_1}\left(\zeta_O \mid \mathbf{x}^0, \zeta_O^0\right)
    =\frac{e^{R_O\left(\mathbf{x}^0, \zeta_E, \zeta_O \right)}}{\sum_{\tilde{\zeta}^j \in \mathbf{D}_O} e^{R_O\left(\mathbf{x}^0, \tilde{\zeta}_E, \tilde{\zeta}^j\right)}}\\
    \end{gathered}
\end{equation}

\textcolor{black}{where $P_{L_1}\left(\zeta_O \mid \mathbf{x}^0, \zeta_O^0\right)$ denotes the probability of the level-1 object choosing trajectory $\zeta_O$. }
Adopting virtual features, the longitudinal interaction can be represented with virtual time headway from the object to the ego. Furthermore, the belief of the ego vehicle lane change preference  $\mu \left(\tilde{\theta}^{md}_E\right)$ can directly influence the reward of the object. 

\begin{equation}
    R_{O}\left(\mathbf{x}^0, \zeta_E, \zeta_O, \tilde{\theta}^{md}_E\right) = 
    \mathbb{E}_{\tilde{\theta}_O^{\text{THWF}}}
    \left(
    \boldsymbol{\theta_{O}}^T \sum_{t=0}^{N-1} \mathbf{f}_{O}\left(\mathbf{x}^t, \zeta_E^t, \zeta_O^t, \tilde{\theta}^{md}_E\right)\right)
\end{equation}

\textcolor{black}{where $\theta_{O}$ and $\mathbf{f}_{O}$ denote the object's preference weights and functional features, respectively; their dimensions are specified in Table \ref*{table:func_ft}
}
The virtual object feature of time headway to the front is modeled with the possible preceding vehicles and the ego. 
\begin{equation}
    f_{risk_f}=e^{\left((p(\zeta'_{LK}))\text{THWF}_O+p(\zeta'_{LC})\text{THWF}'_O\right)}\\
\end{equation}
where $f_{risk_f}$ serves as a dimension of the functional feature $\mathbf{f}_{O}$. $\text{THWF}_O$ denotes the actual time headway to the preceding vehicle in the time horizon given $\zeta_E$, while $\text{THWF}'_O$ is the object's virtual time headway to the ego. The probability of ego lane change $p(\zeta'_{LC})$ and lane keeping $p(\zeta'_{LK})$ in the next horizon is further predicted to determine their weights. When a lane change is more likely to happen in the next step, the virtual time headway to the ego receives a higher weight. 

Naturally, 
\begin{equation}
\begin{gathered}
    p(\zeta'_{LC})+p(\zeta'_{LK})=1
    \end{gathered}
\end{equation}

We approximate the softmax likelihood with a Gaussian likelihood, whose mean is the reward of a certain trajectory. It explains the elements of the one-hot vector by a Gaussian density. This is a trick established earlier \cite{williams2006gaussian, patacchiola2020-bayesianmetalearningfewshot}. 

\begin{equation}
    p(\zeta'_{LC}) \sim \mathcal{N}\left(\boldsymbol{\theta_{E}}^T\mathbf{f}_{E}^{i_2}, \sigma^2\right)
\end{equation}
    
\begin{equation}
\begin{aligned}
    p(\zeta'_{LC})-p(\zeta'_{LK})=&\theta_E^{\text{THWF}}(f_{risk_f}-f_{risk_f}')\\&+
    \theta_E^{\text{THWB}}f_{risk_r}+\theta_E^{md}f_{md}
    \end{aligned}
\end{equation}

where $f_{risk_f}$ and $f_{risk_r}$ denote the predicted ego risk level to the front and back if the ego changes lanes in the next step, with a slight abuse of symbols. Meanwhile, $-f_{risk_f}'$ refers to the risk level to the front if it keeps its lane.


%% file: sections/3-3-L_2_func_rewards.tex
\subsection{Level-2 Game}\label{sec:3-3-bp_ip}

While the object vehicle bases its decisions on functional rewards, the ego vehicle is assumed to determine its behavioral distribution $P\left(\zeta_E \mid \mathbf{x}^0\right)$ according to communication utilities. 

\begin{equation}\label{eq:probs-e}
P\left(\zeta_E \mid \mathbf{x}^0\right)
=\frac{e^{R_E\left(\mathbf{x}^0, {\zeta}_E\right)}}{\sum_{\tilde{\zeta}^i \in \mathbf{D}_E} e^{R_E\left(\mathbf{x}^0, \tilde{\zeta}^i\right)}}
\end{equation}

where $R_E\left(\mathbf{x}^0, \zeta_E \right)$ denotes the ego vehicle's total social compatibility reward, $\tilde{\zeta}^i$ is a generated trajectory of the object that has the same initial state as $\zeta_E$, $\mathbf{D}_E$ is the generated trajectory set. 

$R_E$ is determined by the communicative features and weights $\boldsymbol{\theta_{sc}}, {\mathbf{F}_{\zeta_E}}$. 

\begin{equation}\label{eq:thetaF}
R_{E}\left(\mathbf{x}^0, \zeta_E\right) = 
\mathbb{E}_{\tilde{\theta}^{md}_E \sim \mu}
\boldsymbol{\theta_{sc}}^T \sum_{t=0}^{N-1} \mathbf{F}_{\zeta_E}\left(\mathbf{x}^t, \zeta_E^t\right)
\end{equation}

where \(\mathbf{F}_{\zeta_E}  \in {R^4}\) denotes four higher-level social-compatible utilities. \(\boldsymbol{\theta_{sc}}  \in {R^4}\) denotes their weights. 


\begin{equation}\begin{split}
\mathbf{F}_{\zeta_E}\left(\mathbf{x}^t, \zeta_E^t\right)=
&\left[
R_{E_{\text {SI}}}\left(\mathbf{x}^0, \zeta_E\right) 
 , R_{E_{\text {pride}}}\left(\mathbf{x}^0, \zeta_E\right), \right.\\
&\left. R_{E_{\text {inquiry }}}\left(\mathbf{x}^0, \zeta_E\right), R_{E_{\text {benef }}}\left(\mathbf{x}^0, \zeta_E\right)
\right]
\end{split}
\end{equation}

where the ${R_{E_{\text {SI}}}}$, ${R_{E_{\text {pride}}}}$, ${R_{E_{\text {inquiry }}}}$, and ${R_{E_{\text {benef }}}}$ are respectively the ego vehicle's self-interest utility, pride utility, inquiry utility, and beneficence utility. 


They can be categorized into two main sections: egoism and altruism. Both of these categories comprise two classes of utilities known as functional and communication utilities. 

Functional utilities refer to the anticipated benefits that result from the interaction of vehicles, which are determined by their characteristics in terms of safety, efficiency, comfort, and energy conservation.
They describe the drivers' expected rewards, whose uncertainty is mainly caused by others' uncertain behavior choices. 

The functional utility of an ego agent can refer to Eq. (\ref*{eq:6}) and be calculated considering the behavior distribution $P_{L_1}\left(\zeta_O \mid \mathbf{x}^0, \zeta_E\right)$ of the interacting object vehicle.

\textcolor{black}{ In this work, we adopt a level-k hierarchy where the uncertainty of cognitive level is not considered. This is because we do not model the object directly; instead, we model the ego's projection of how it perceives the object. In this sense, a mutual communication model requires the object to be at least level-1 so that it can listen. This makes the ego a level-2 reasoner. Meanwhile, empirically level-3 reasoners are rarely encountered \cite{costa-gomes2009-comparingmodelsstrategic}. 
 }

%% file: sections/3-4-info_rewards.tex
\section{Communication Utility}
\label{sec3_2_C}

Besides the functional rewards mentioned above, the symbolic or communicative value is conveyed by the interactive act itself. This value not only communicates regard and respect for the other and the relationship but also reduces the uncertainty of the interaction to convey predictability and trustworthiness \cite{molm2007-valuereciprocity}. 





In this study, we adopt it to encompass two key elements: pride and prejudice, which describe the mutual information flow between the interacting vehicles. 
In a lane-changing context, pride describes the ego expression process and indicates how determined the ego is to change lanes, while prejudice describes the object expression process and its determination not to yield. 

Further, inquiry is proposed for expected prejudice correction in communication. It measures the ego's ability to actively probe for knowledge about the object's perseverance. 

\subsection{Pride and Prejudice}
\label{stereotype}

Pride and prejudice are proposed to describe how clearly information is expressed through information gain.
Information gain is calculated with the Kullback-Leibler (KL) divergence \cite{kullback1997information}. 
Here, we consider communication as a way to correct prejudice and exude pride with clarity. Clarity is defined as the information gain caused by expression in communication.  


\subsubsection{Prejudice} 
\label{ssub:prejudice}

First, prejudice is defined as follows. 
\begin{equation}
    R_{preju}=D_{K L}
            \left[P^*\left(\theta_O\right)
            \| P_{0}(\theta_O)
            \right]
    \label{eq:preju}
\end{equation}

\textcolor{black}{where $P^*$ denotes the attribute probability and $P_0$ the prior belief distribution. }
The ego prejudice about the object can be approximated as the object's clarity. 
\begin{equation}
    R_{O_{clarity}}=D_{K L}
                \left[P_{L_1}\left(\theta_O \mid \mathbf{x}^0, \zeta_O\right)
                \| P_{0}(\theta_O)
                \right]
    \label{eq:obj_clar}
\end{equation}

In this sense, listening corrects prejudice and changes the ego's own mind. Conversely, if bullying AVs is understood as stereotyping the AVs with discrimination, it could be corrected with implicit communication.

\subsubsection{Pride} 
\label{ssub:pride}

We further propose the algorithmic pride as the presumed object prejudice about the ego, which is the ego prejudice over the second-order belief. 
\begin{equation}
    R_{pride}=D_{K L}
            \left[P^*\left(\theta_E\right)
            \| P_{0}(\theta_E)
            \right]
    \label{eq:pride}
\end{equation}
Pride is approximated with ego clarity. 
\begin{equation}
    R'_{E_{clarity}}=D_{K L}
            \left[P_{L_0}\left(\theta_E \mid \mathbf{x}^0, \zeta_E\right)
            \| P_{0}(\theta_E)
            \right]
   \label{eq:ego_clar}
\end{equation}
In this sense, expression exudes pride and tells others what to think. 
Abnormal defensive deceleration can be interpreted as an exudation of pride, related to an indifference to the beliefs of others, resulting in a lack of situational awareness and potentially dangerous behavior towards vehicles following behind.

Specifically, we propose to define pride as the object information gain of ego mandatory lane change preference $\theta_E^{md}$ given the ego action. 

We adopt the following expression for the pride reward. 

\begin{equation}
    \begin{split}
        R_{E_{pride}}\left(\mathbf{x}^0, {\zeta}_E\right) 
            =&e^{\mathbb{E}_{\tilde{\zeta}^i }
            \left[C_{EX-\tilde{\zeta}^i}\right]}\\
            &\left(1-e^{-D_{K L}
            \left[P_{L_0}\left(\theta_E \mid \mathbf{x}^0, \zeta_E\right)
            \| P_{0}(\theta_E)
            \right]}\right).
    \end{split}
\label{eq:sc pride}
\end{equation}

Here, the exponent has two components. On one hand, the KL divergence measures the distance between the posterior and prior distributions of object belief over ego preference.
On the other hand, the $C_{EX-\tilde{\zeta}^j}$ denotes the expression condition factor as a consequence of the trajectory $\tilde{\zeta}^i$. $C_{EX-\tilde{\zeta}^j}$ describes the value of the trajectory terminal state, which will be elaborated in \ref{ea}. 
$M$ denotes the total number of generated ego trajectories. 

\subsection{Inquiry} Inquiry is the act of probing for mutual information. 
 Mutual information $I(X; Y)$ is the reduction in the uncertainty of X due to the knowledge of Y \cite{cover1999elements}. The ability to acquire knowledge through active inquiry is modeled as maximizing the mutual information between the $\tilde{\theta}_O^{\text{THWF}}$ and the $\zeta_O$ given the $\tilde{\theta}_E^{md}$.


\begin{equation}
    \begin{split}
        R_{E_{inquiry}}\left(\mathbf{x}^0, {\zeta}_E\right)
        =&e^{\mathbb{E}_{\tilde{\zeta}^j
        }\left[C_{LI-\tilde{\zeta}^j}\right]
        }
        \mathbb{E}_{\theta_E^{md}}
        \left(
        I(\zeta_O;\tilde{\theta}_O^{\text{THWF}}|\tilde{\theta}_E^{md})
        \right)
    \end{split}
    \label{eq:sc inqui}
    \end{equation}

    where $C_{LI-\tilde{\zeta}^j}$ denotes the active learning condition as the consequence of the trajectory $\tilde{\zeta}^j$, which will be elaborated in \ref{if}. $I(\zeta_O;\tilde{\theta}_O^{\text{THWF}}|\tilde{\theta}_E^{md})$ is computed with distance of the object car's real behavior distribution $P\left(\zeta_O \mid \mathbf{x}^0, \zeta_E, \theta_O\right)$ from $P_0\left(\zeta_O \mid \zeta_E\right) $, given the ego's prior belief. 

    \begin{equation}
        \begin{split}
        I
        =
        &\mathbb{E}_{\tilde{\theta}_O^{\text{THWF}}}
        \left(
        D_{K L}
        \left[P_{L_1}\left(\zeta_O \mid \tilde{\theta}_E^{md}, \tilde{\theta}_O^{\text{THWF}}\right)
         \| P_{L_1}\left(\zeta_O \mid \tilde{\theta}_E^{md}\right)
         \right] \right)
         \\=&
         \sum_{q=1}^{N_O}P_0(\tilde{\theta}_O^{\text{THWF}})
         \sum_{j=1}^M
         P_{\tilde{\zeta}^j \mid \theta_O^q }
         \log \frac{P_{\tilde{\zeta}^j\mid \theta_O^q }}
         {\sum_{q=1}^{N_O}P_0(\tilde{\theta}_O^{\text{THWF}})P_{\tilde{\zeta}^j_{0}}}.
    \end{split}
\label{eq:sc info-inqui}
\end{equation}

\textcolor{black}{ This treatment of epistemic uncertainty in mutual communication helps mitigate the structural bias introduced by the cognitive model. A pure Stackelberg game may induce aggressive behavior because it assumes that the object only reacts. In the mutual communication framework, however, the ego optimizes not only expression through its own action but also the listening channel, which counterbalances that bias. Therefore, the mutual communication model supports a more reciprocal interactive decision process.
}

Additionally, inquiry can be proven to be the expected prejudice correction over the object actions. Therefore, it is adopted as the measurement of epistemic uncertainty. It can be reduced with stereotyping if we define stereotyping as the entropy of the prior belief. Details can be found in Appendix A. 

\subsection{Communication Affordances}
\label{niche}

Based on communication rewards, communication affordance models are proposed to describe the need for the ego or object to listen to each other. This concept was coined to describe the attention paid to environmental information \cite{gibson1977theory}. Later, the role of social cognition was discussed for the perception of social affordances \cite{fiebich2014-perceivingaffordancessocial}. In this section, communication affordances for expression and listening are modeled with the safety and incentive conditions for lane-changing decisions. 
\subsubsection{Safety Condition}
\label{sf}
\begin{figure}
    \centering
    \includegraphics[scale=0.4, trim=150 150 280 170, clip]{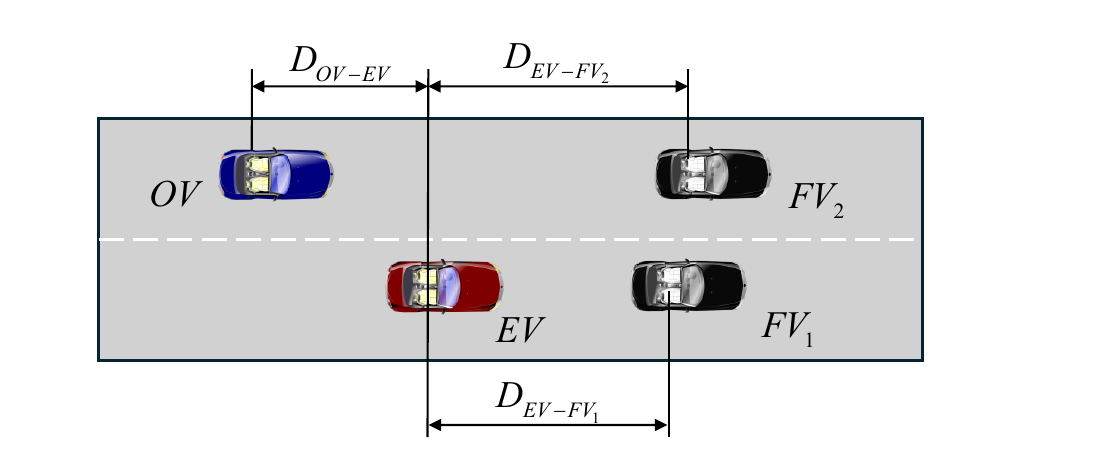}
    \caption{Layout of the lane change process for the probabilistic model of niches}
    \label{fig:layout}
\end{figure}

The safety conditions are defined with an adapted probabilistic condition model from \cite{cui2023-passingyieldingintentionestimation}. Here, conditions reflect lane-changing safety at the end of the chosen trajectory. There are two metrics encompassing ego vehicle lane change safety conditions, the car following condition $ C_{CF}$  and lane change condition $C_{LC}$. The first considers the longitudinal distance to the preceding and following vehicles in the target lane $D_{OV-EV}$ and $ D_{EV-FV}$ to assess the risk as shown in Fig. \ref*{fig:layout}. The $C_{LC}$ checks how safe the ego is to pass the object and travel the lateral distance to the target lane. 

\subsubsection{Incentive Condition}
\label{if}

The incentive condition has a negative influence on courtesy. 
Therefore, we adopt the following model of incentive condition $C_{IF}$ inspired by the IDM \cite{PhysRevE.62.1805}.

\begin{equation}
    \begin{aligned}
        C_{IF} &= \min\{1, \frac{v_{LK}}{v_{LC}}\}\\
        v_{LK} &= \min\{v_{\zeta_E}, v_{FV_1}\}\\
        v_{LC} &= \min\{v_0, v_{FV_2}\}
    \end{aligned}
\end{equation}

where $v_{LC}$ represents the anticipated speed after changing lanes, while $v_{LK}$ signifies the projected speed if no lane change occurs. $v_{\zeta_E}$ is the terminal ego velocity of a trajectory $\zeta_E$, while $v_0$ signifies a desired velocity, which is 20 m/s here. The upper limit of the variable $C_{IF}$ is established at 1 to maintain compatibility with discretionary lane changes.

\subsubsection{Expression Affordance}
\label{ea}
The expression affordance $C_{EX}$ describes how much the object would like to listen at the end of the trajectory selected. It integrates the two parts of safety conditions $C_{CF}$ and $C_{LC}$ differently. 

\begin{equation}
C_{EX} = C_{CF}+ C_{LC}
\end{equation}

which indicates that the ego speaks louder with a better chance of a successful lane change, whether longitudinally or laterally. 

\subsubsection{Listening Affordance}
\label{la}
The listening affordance $C_{LI}$ describes how much the ego needs to listen. The model is also constructed with safety parameters $C_{CF}$ and $C_{LC}$, while being influenced by the incentive factor as well.

\begin{equation}
    C_{LI} = C_{IF} (C_{CF}- C_{LC})
    \end{equation}

First, with more desperation for incentives comes less listening motive. Second, the longitudinal advantage leads to better listening quality, while lateral achievement may lead to a certain future with less need to listen.

\subsection{Pride-Inquiry (P-I) Plane and the Pride-Prejudice (P-P) Plane}
\label{plane}

In order to describe the intensity and tendency of one-sided engagement and mutual engagement, we integrate three factors to form two planes, respectively. Ego pride and ego inquiry are adopted to form the Pride-Inquiry (P-I) plane for expression and listening. Meanwhile, ego pride and prejudice form the Pride-Prejudice (P-P) plane for bi-directional expressions. 

According to Eq. (\ref*{eq:sc pride}), pride experiences a normalization procedure. If we set conditions to be 0 to eliminate the influence of future conditions, it can be guaranteed that the resulting pride value falls within the range of [0, 1]. 
This is the same with the corrected ego prejudice, or rather, the exuded object pride. 
The ego inquiry is not normalized, yet it is nonnegative and usually smaller than 1. 

\begin{figure}
    \centering
    \includegraphics[scale=0.4, trim=150 10 80 50, clip]{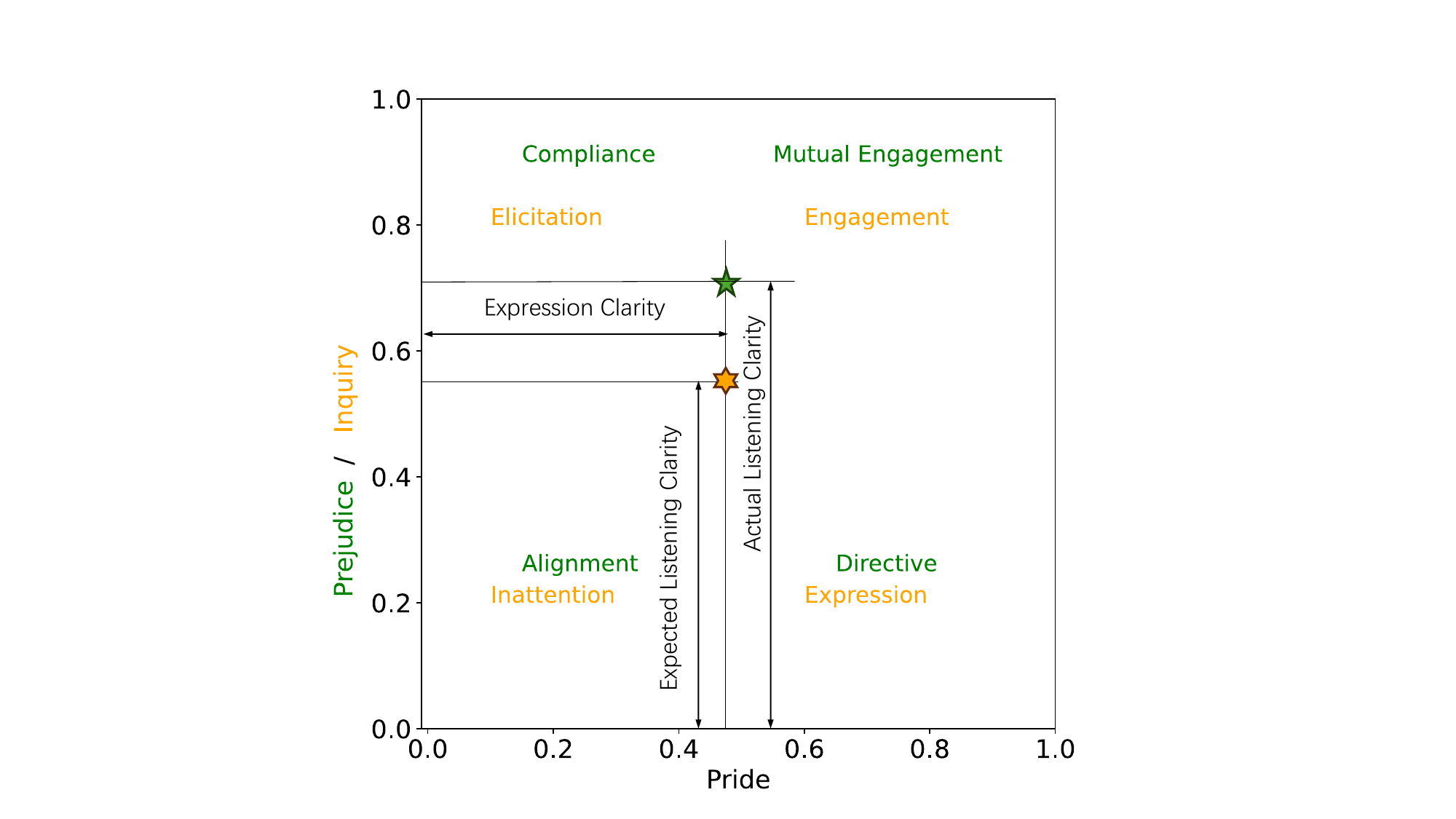}
    \caption{The Pride-Inquiry (P-I) plane and Pride-Prejudice (P-P) plane to demonstrate the
    expression clarity, expected and actual listening clarity of the interacting vehicles}
    \label{fig:plane}
\end{figure}

Combined together, pride rewards and inquiry rewards form a P-I plane to describe the state of the ego vehicle in communication. 
If only one of the communication rewards is high and the other is low, the ego would be in the state of expression for pride or elicitation for inquiry. This diagonal line describes the ego tendency of listening or expression. 
The ego engages in interactive communication through pride and inquiry. However, when the rewards of communication are low, the ego may become inattentive. Therefore, the diagonal line of the P-I plane visualizes levels of engagement.

Mutual expressions define the P-P plane in mutual communication. 
An increase in prejudice paired with lower pride results in ego compliance, whereas the opposite suggests ego directive. This diagonal line serves as a visual representation of mutual communication tendencies. When there is heightened clarity on both ends, it signifies active engagement in communication, reflecting a high level of involvement. Through the reciprocal expression, communication sublates both pride and prejudice, ultimately aligning both parties.

\textcolor{black}{
Specifically, unlike their counterpart in aleatoric uncertainty modeling \cite{11048679}, which captures single-agent control, our communication planes illustrate both the mutual information of the listening channel determined by the ego and the manner in which object actions govern prejudice disclosure. 
The Pride-Inquiry and Pride-Prejudice planes thus capture the object's impact on interaction mechanisms and uncertainty results, reflecting the bidirectional nature of mutual communication.}

%% file: sections/4-0-mirl.tex
\label{sub:MIRL}

To quantitatively model cognitive uncertainty in communicative driving scenarios, we propose a Communication-based Multi-Agent Inverse Reinforcement Learning (C-MIRL) approach adapted from our former work \cite{11048679}. 

The input to our framework consists of two complementary data sources, NGSIM dataset and DIL experiments, to ensure both ecological validity and controlled validation. 
\textcolor{black}{
While NGSIM captures human-human interactions rather than AV-human interactions, the communication rewards we calibrate (pride, prejudice, inquiry) capture behavioral patterns that transfer across interaction types. 
The DIL experiment provides empirical evidence that these patterns generalize to AV-human scenarios, with positive correlations between model predictions and human evaluations.}

%% file: sections/5-1-data-2-lf.tex
Every interaction scene extracted from NGSIM consists of $35 \times 33 \times 33 = 38115$ generated 3-s trajectories multiplied by 6 virtual feature trajectories for each $\theta_E^{md}$ histogram. Four histograms ranging from $R_{min}=-4$ to $R_{max}=0$ are adopted for the belief updating of the ego over $\theta_O^{THWF}$, while 6 histograms ranging from $R_{min}=0$ to $R_{max}=9$ are adopted for the belief of the object over $\theta_E^{md}$. Training usually takes 18 s for each learning epoch with an Intel Xeon Processor E5-2678 v3 CPU, resulting in about 2.5 h for a 500-iteration training process. 

%% file: sections/5-exp-res.tex

\subsection{Case Analysis}
\label{sec4_D}

We analyze representative lane-changing scenes to reproduce the interactions and interpret the learned rewards.

\textcolor{black}{Each interaction pair is divided into 50 three-second scenes for open-loop rolling-horizon planning \cite{huang2022-drivingbehaviormodeling}. We use 35 scenes for training and 15 for testing. The original lane-changing processes or attempts typically lasted about 15 seconds, which includes the proactive communication process and provides enough interaction context for personalized driver modeling. The division into 50 scenes provides adequate temporal resolution together with sufficient training and testing samples. The extracted scenes were sampled as 3-second intervals for planning to account for interactions in the dynamic game. }

\input{sections/5-2_inquiry_case.tex}

This case illustrates how the proposed communication rewards improve calibration.

\input{sections/5-4_cmn_graph_case.tex}
\input{sections/5-5_stats_analys.tex}
\input{sections/5-6_comparison.tex}
\input{sections/5-7_ablation_study.tex}
\input{sections/5-8-HIL.tex}

%% file: sections/5-2_inquiry_case.tex
\subsubsection{Communication Utility of Inquiry}

\begin{figure}[ht]
    \centering
    \label{fig:358 scene}
    \begin{minipage}[b]{\linewidth}
    \centering
        \begin{minipage}[b]{0.49\linewidth}
            \subcaptionbox{\label{fig:358 scene-curve}}{
            \includegraphics[width=\linewidth]{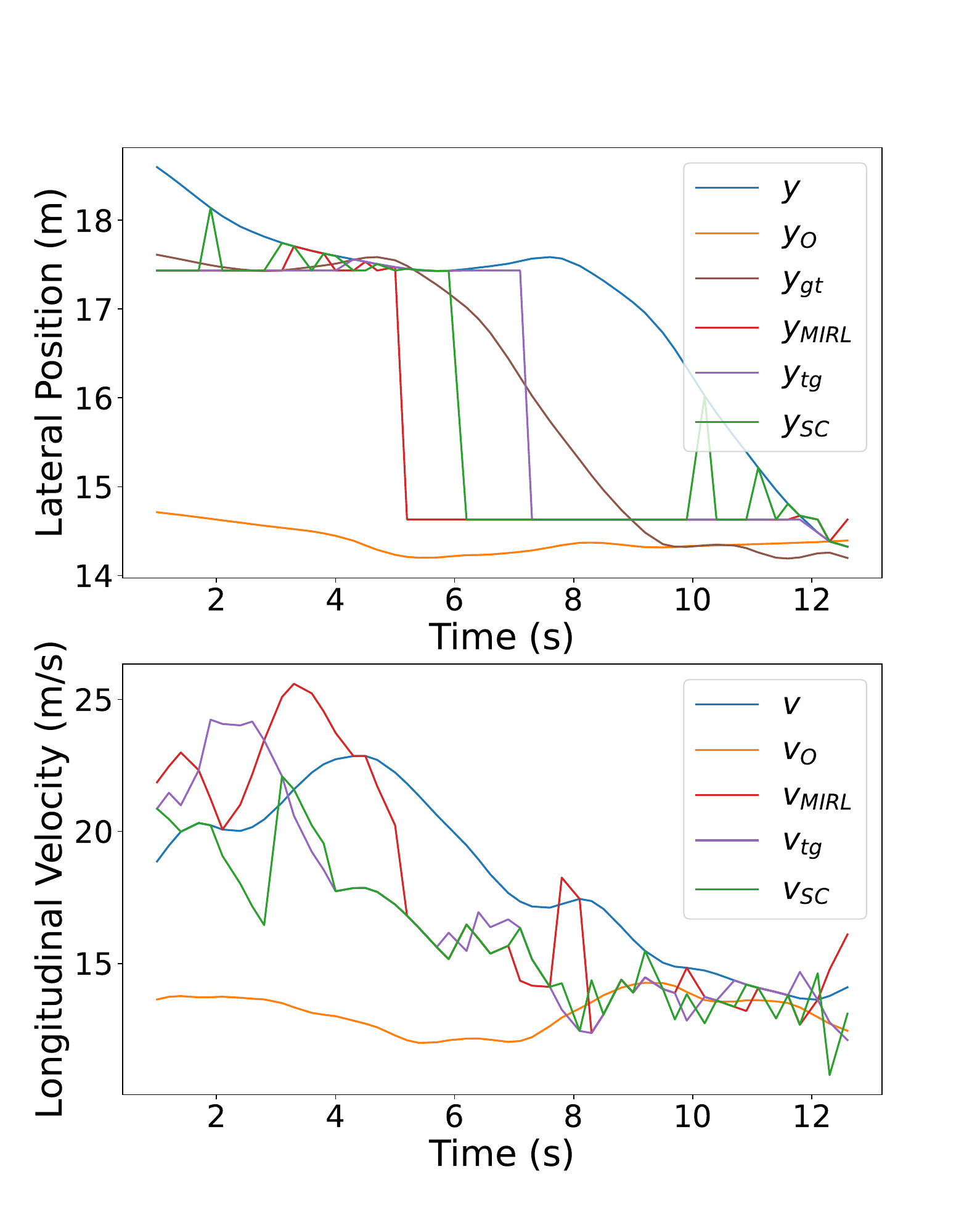}
        }\end{minipage}
        \begin{minipage}[b]{0.49\linewidth}
            \subcaptionbox{\label{fig:195 E}}{
            \includegraphics[width=\linewidth]{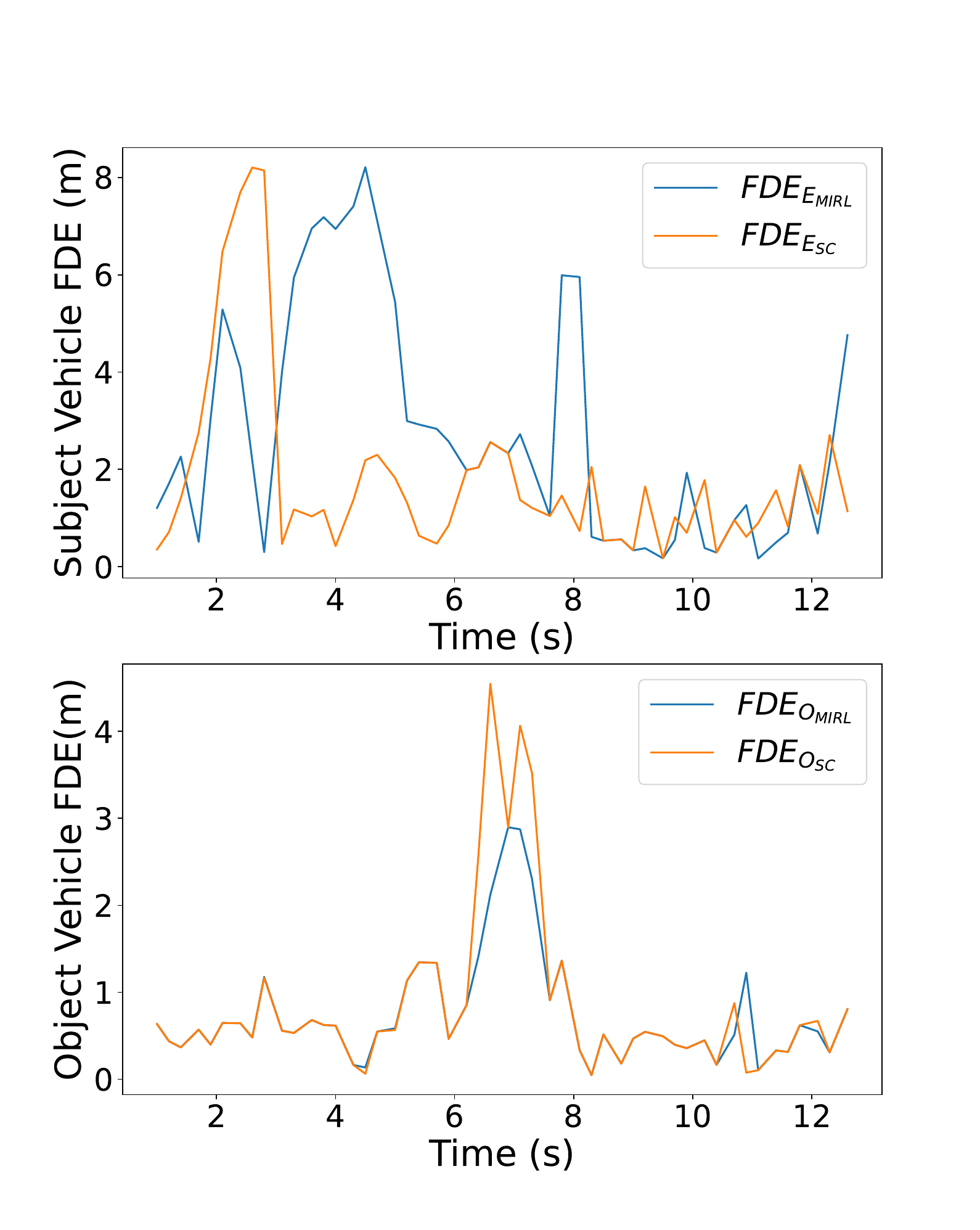}
        }\end{minipage}
        \end{minipage}
        \caption{Comparison of calibrated model behavior for vehicle 195 with C-MIRL and MIRL. (a) Lateral position and speed curves during the interaction process. $y$ and $v$ denote lateral position and velocity, respectively. $y_O$ and $v_O$ are the object curves, $y_{gt}$ is the 3-second ground-truth lateral position, and $y_{tg}$ and $v_{tg}$ are the sampled target values. $y_{MIRL}$ and $v_{MIRL}$, and $y_{C}$ and $v_{C}$ are the calibrated model behaviors from MIRL and C-MIRL, respectively. (b) The best-of-3 FDE of the ego vehicle and the object calibrated by C-MIRL and MIRL.  
        }
    \end{figure}

\begin{figure}
\begin{subfigure}[b]{\linewidth}
\centering
\includegraphics[scale=0.22, trim=35 0 0 10, clip]{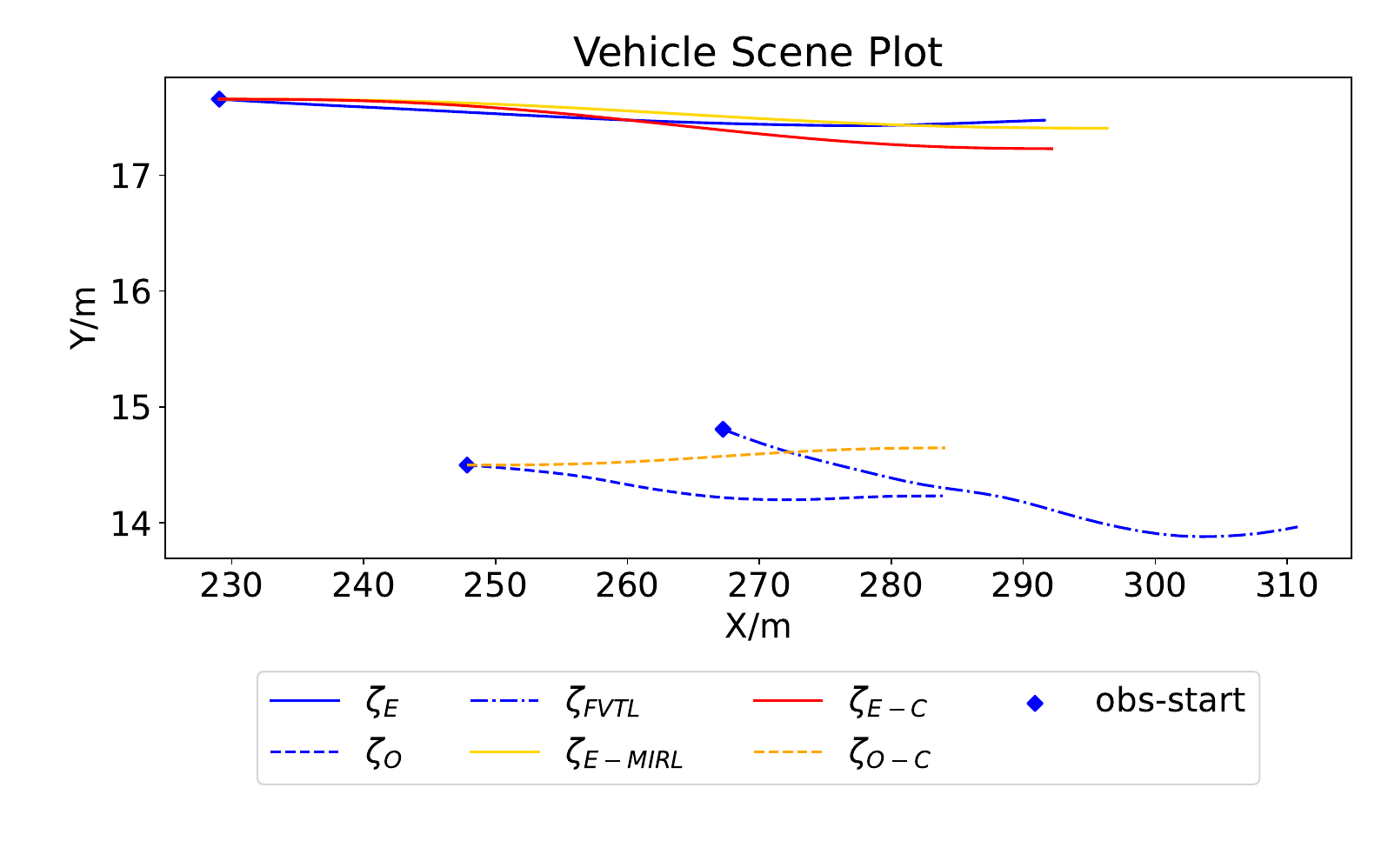}
\caption{}
\label{fig:195 subfig:a}
\end{subfigure}
    \begin{minipage}[b]{\linewidth}
     \begin{minipage}[b]{0.45\linewidth}
     \subcaptionbox{\label{fig:195 36 clar}}{
            \includegraphics[width=\linewidth]{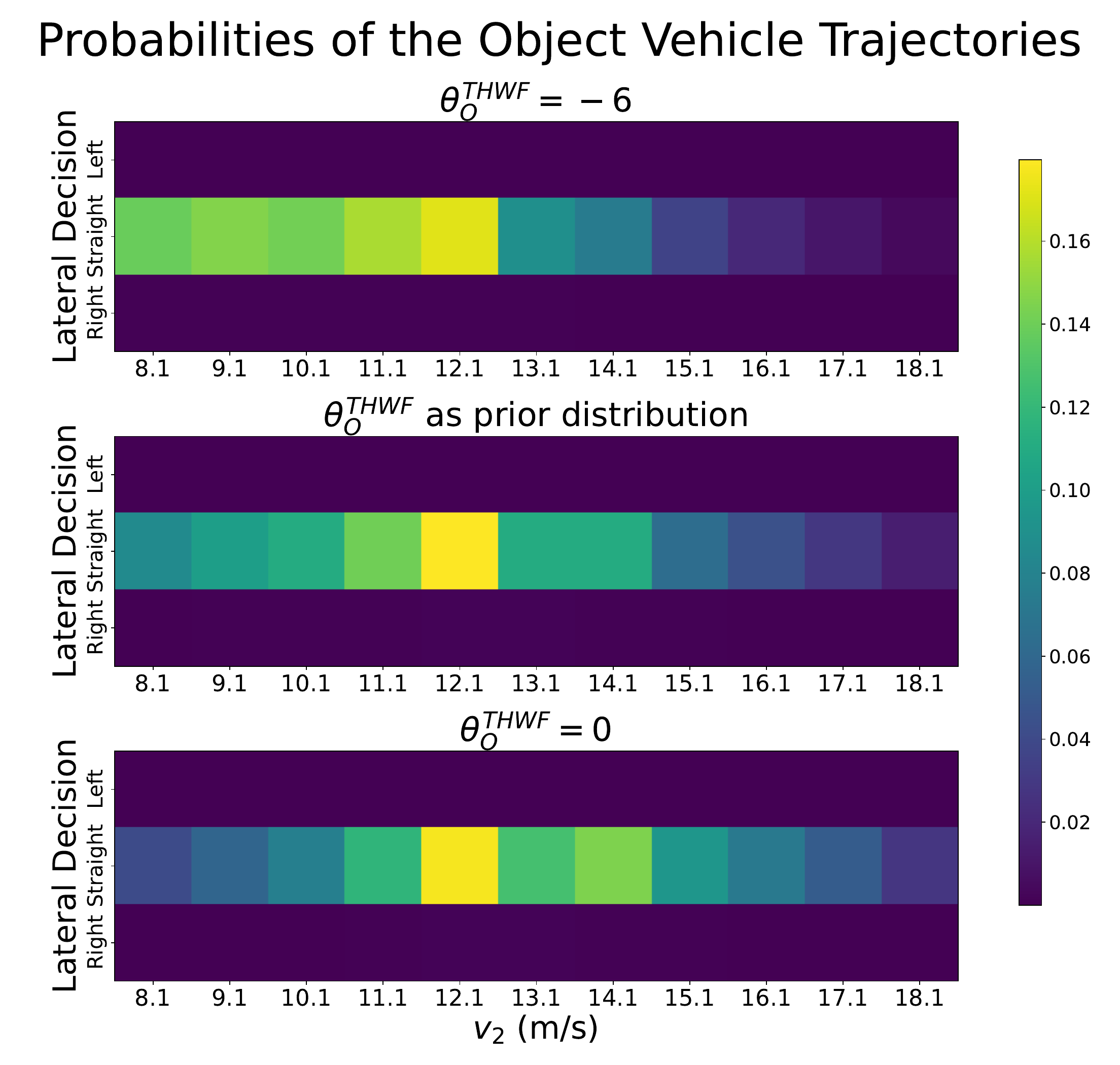}
            }\end{minipage}
    \begin{minipage}[b]{0.45\linewidth}
    \subcaptionbox{\label{fig:195 36 inqr}}{
\includegraphics[width=\linewidth]{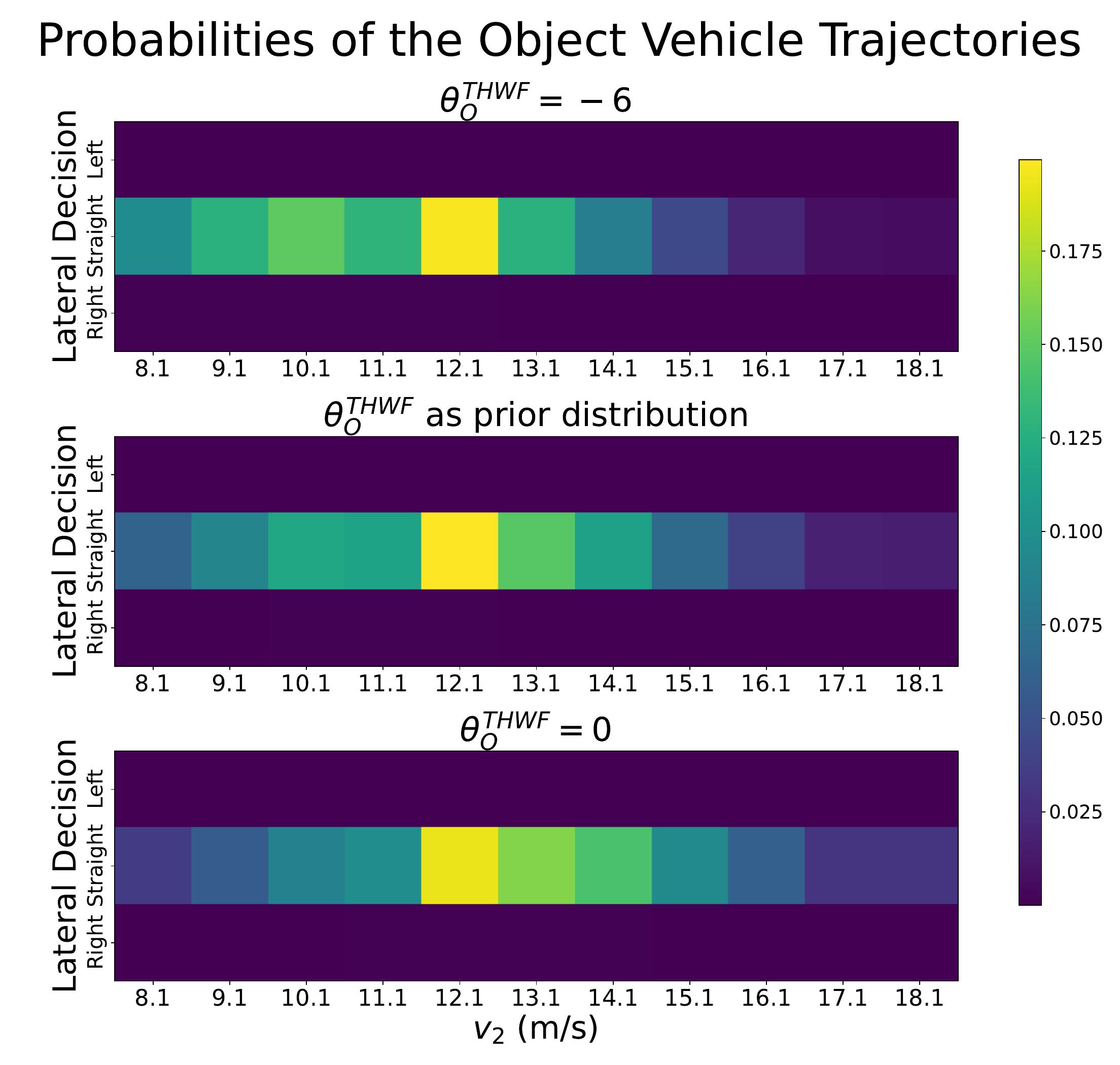}
}\end{minipage}
\end{minipage}
    \caption[short]{The possibility distributions
over the actions of the object
vehicle given actions of the ego at 3.6 s. (a) The scene that starts at 3.6 s. obs-start denotes the observation at 3.6 s. The blue lines denote the ground truth of the ego vehicle $\zeta_E$, the object $\zeta_O$, and the preceding vehicle in the target lane (FVTL). The calibration results of the ego and the object from C-MIRL are denoted by $\zeta_{E-C}$ and $\zeta_{O-C}$, respectively. (b)(c) The object behavior distribution under different $\theta^{THWF}_O$ assumptions at 3.6 s, given different ego actions selected by C-MIRL and MIRL, with the perceived $\theta_E^{md}$ as 5.4.}
    \label{fig:6 195}
\end{figure}

    Fig.\ref*{fig:358 scene-curve} shows vehicle 195 changing lanes in front of vehicle 194 in NGSIM US-101. C-MIRL captures the early deceleration and reduces ego FDE by up to 6 m after 3 s, as shown in Fig.\ref*{fig:195 E}. Fig.~\ref*{fig:195 subfig:a} presents the candidate and ground-truth trajectories together with the interaction context. 

At 3.6 s, the initial speed is $v_0=22.23$ m/s. C-MIRL selects lane-change anticipation with deceleration to about $v_0-2$ m/s, close to the human action, whereas MIRL keeps the lane and accelerates. Tables \ref*{table:3} and \ref*{table:4} show that MIRL prefers lower jerk and velocity cost, while C-MIRL accepts a lower self-interest reward because the inquiry reward is substantially higher.

\begin{table*}[h!t]
\centering
\begin{tabular}{ccccccccccc}
   \toprule
    algo  &traj & $R_v$ & $R_{a_x}$ & $R_{a_y}$ & $R_{\text{jerk}}$ &$R_{\text{ THWF}}$ & $R_{\text{THWB}}$ & $R_{\text{colli}}$ & $R_{\text{Social Impact}}$ & $R_{md}$ \\
   \midrule
   C-MIRL  &LA($v_0-3$)&\textbf{2.595}& 0.071& 0.06& \textbf{-1.161}& -0.0& -0.719& -0.001& -0.163& 0.028\\
   MIRL  &LK($v_0-1$) &2.86& 0.075& 0.023& -0.741& -0.0& -0.506& -0.002& -0.161& 0.015 \\
   \bottomrule
\end{tabular}
\caption{Weighted functional rewards of vehicle 195 at scene 3.6 s, assuming the actions chosen by C-MIRL and MIRL. }
\label{table:3}
\end{table*}

\begin{table}[h!t]
\centering
\begin{tabular}{cccccc}
   \toprule
   algo  &traj  & $R_{E_{SI}}$ & $R_{E_{\text{pride}}}$  & $R_{E_{\text {inquiry }}}$  & $R_{E_{\text {benef}}}$\\
   \midrule
   C-MIRL  &LA($v_0-2$)  &4.064& 0.0& \textbf{1.783}& -0.288\\
   MIRL&LK($v_0+3$)  & 4.176& 0.0& 0.911& -0.314\\
   \bottomrule
\end{tabular}
\caption{Weighted social compatibility rewards of vehicle 195 at scene 3.6 s, assuming the actions chosen by C-MIRL and MIRL. }
\label{table:4}
\end{table}

Fig.~\ref*{fig:195 subfig:a} compares the trajectories chosen by the two models. C-MIRL selects lane-change anticipation and deceleration with a higher inquiry reward of 1.783. Fig.\ref*{fig:195 36 clar} shows the object trajectory distribution under different $\theta_O^\text{THWF}$ assumptions: as $\theta_O^\text{THWF}$ approaches 0, the object decelerates less and may even accelerate to block the ego. Compared with the MIRL choice in Fig.~\ref*{fig:195 36 inqr}, the tactful C-MIRL action elicits more informative responses. In particular, Fig.\ref*{fig:195 36 clar} shows that, when $\theta_O^\text{THWF}=0$, the object exudes such a high level of pride that it is prone to accelerating and obstructing the ego rather than maintaining a steady velocity. This illustrates the role of ego inquiry in the Bayesian persuasion game.

%% file: sections/5-4_cmn_graph_case.tex
\subsubsection{Pride-Inquiry (P-I) Plane and Pride-Prejudice (P-P) Plane for Mutual Communication}

\begin{figure}
    \centering
    \begin{minipage}[b]{0.45\linewidth}
        \subcaptionbox{\label{fig:631 Rci plane 2d}}{
        \includegraphics[width=\linewidth]{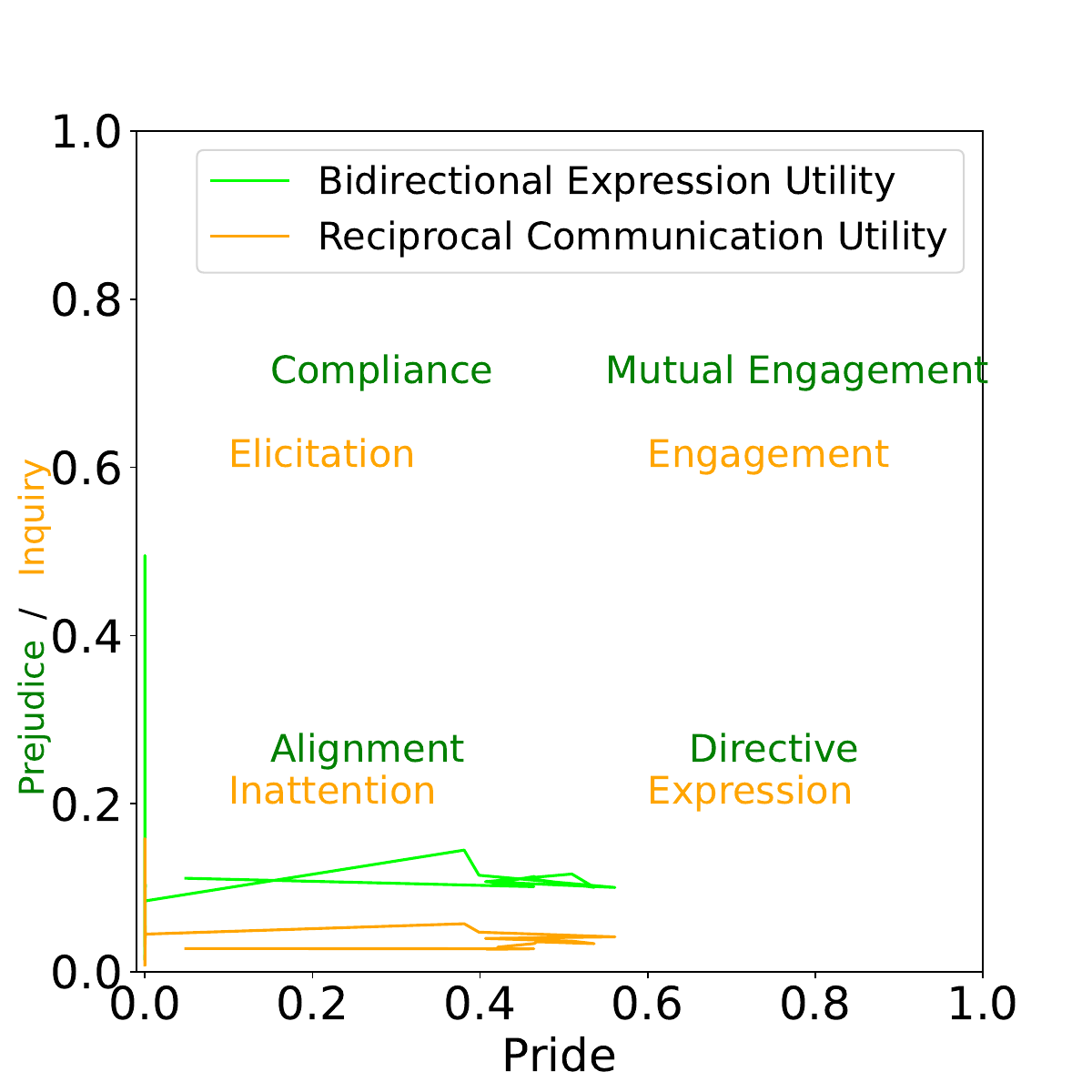}
        }\end{minipage}
    \begin{minipage}[b]{0.45\linewidth}  
        \subcaptionbox{\label{fig:631 Rci plane 3d}}{  
        \begin{minipage}[b]{\linewidth}
            \includegraphics[width=\linewidth]{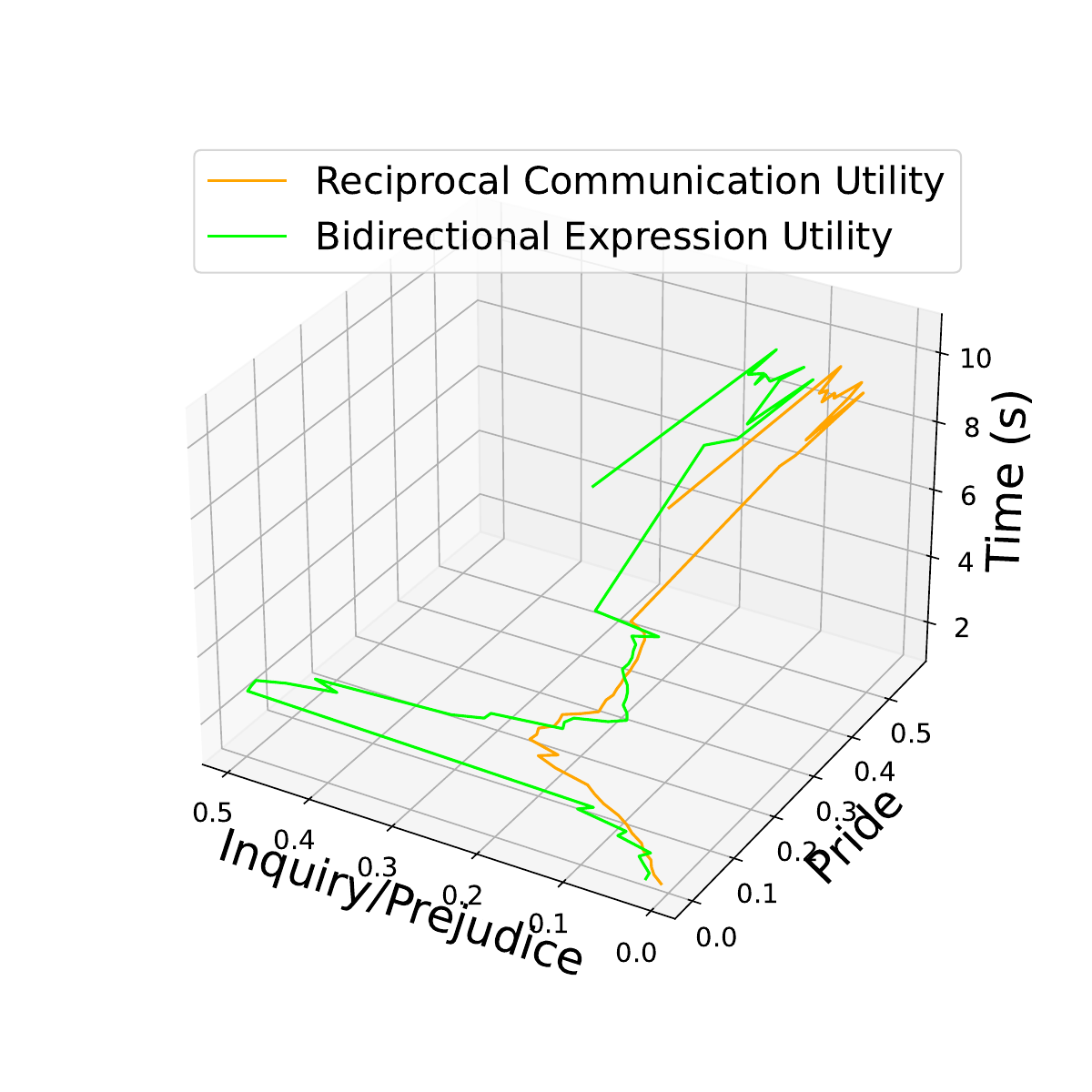}
        \end{minipage}
        }
    \end{minipage}
    \vspace{0.5cm} 
        \begin{minipage}[b]{0.45\linewidth}
        \subcaptionbox{\label{fig:631 reci R_real}}{
        \includegraphics[width=\linewidth]{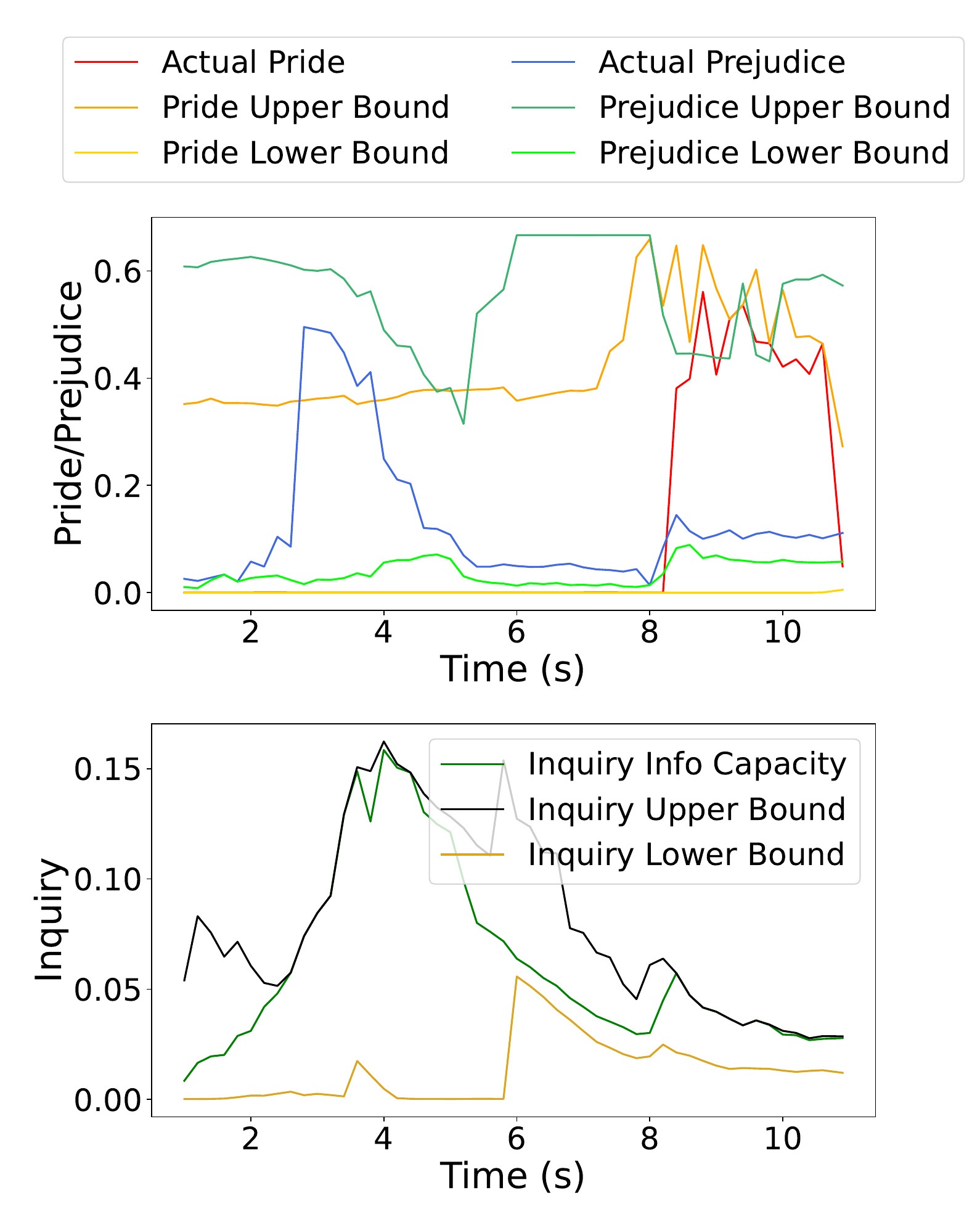}
    }\end{minipage}
       \begin{minipage}[b]{0.45\linewidth}
       \subcaptionbox{\label{fig:631 scene}}{
       \includegraphics[width=\linewidth]{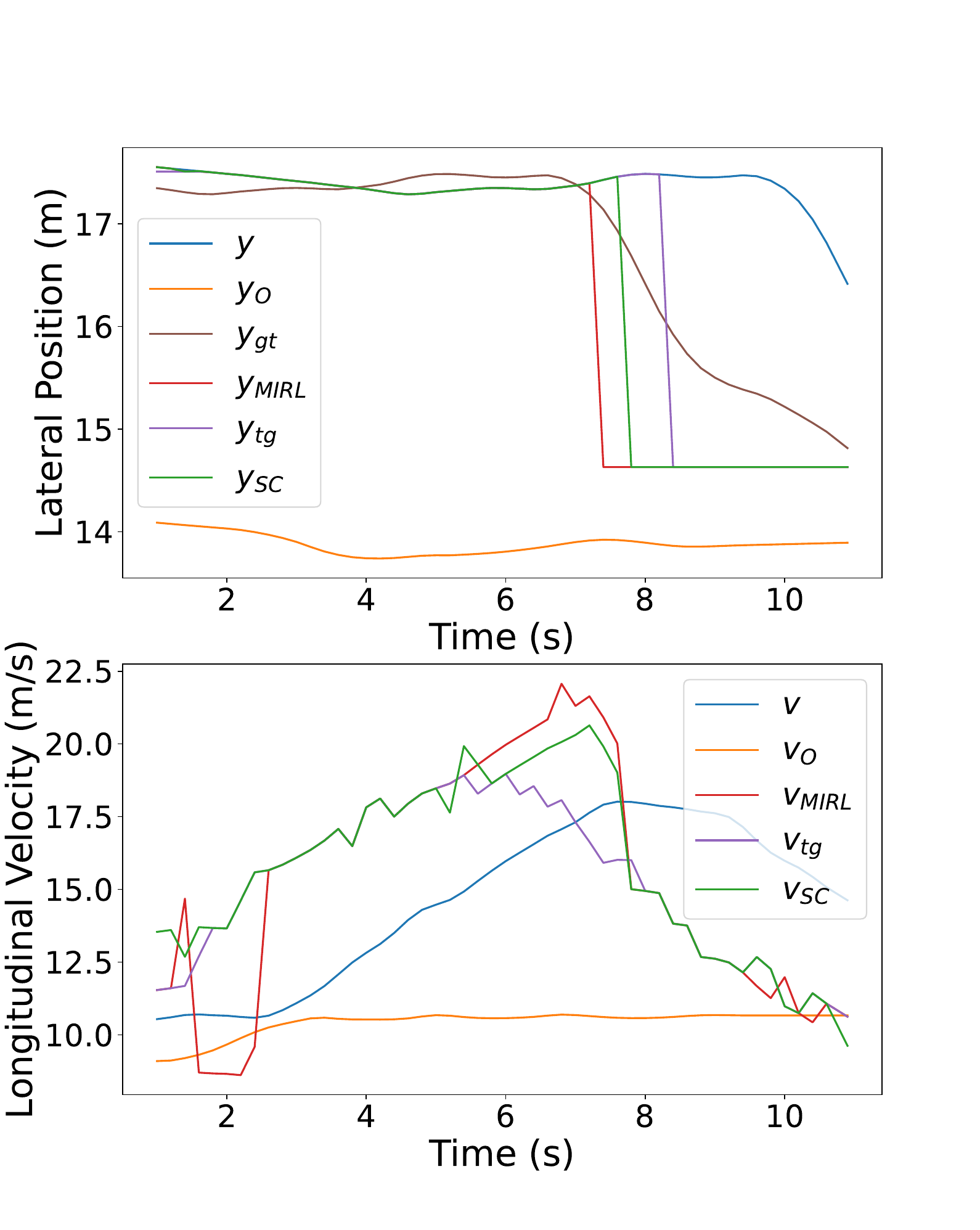}
   }\end{minipage}
    \caption[short]{Interaction intensity and tendency analysis of vehicle 500 for ego engagement and mutual engagement in communication with the P-I and P-P planes. (a) Communication state trajectories for the P-I and P-P planes. (b) Communication state trajectory over time. (c) Communication utility states of pride and inquiry. (d) Interacting vehicle behavior during the lane-change process, including lateral position and speed curves. }
    \label{sr-c-case}
\end{figure}

The two distinct communication planes, namely the Pride-Inquiry (P-I) plane and the Pride-Prejudice (P-P) plane, have been proposed to illustrate the levels of interaction intensity and the inclination for ego engagement and mutual engagement in communication. 
A mandatory lane-changing interaction is examined using the proposed communication planes. The actual reciprocal communication utility and its upper and lower bounds among all possible actions are depicted in Fig.~\ref*{fig:631 Rci plane 2d}, while their temporal evolution is shown in Fig.~\ref*{fig:631 Rci plane 3d} and Fig.~\ref*{fig:631 reci R_real}. 

Three dimensions of the communication plane, the pride, prejudice, and inquiry utilities, are shown in Fig. \ref*{fig:631 reci R_real}. 

Fig.\ref*{fig:631 Rci plane 2d} and Fig.\ref*{fig:631 Rci plane 3d} show that prejudice first increases until 3 s and moves the ego toward compliance in the P-P plane, which could be attributed to its acceleration. At the same time, the ego increases inquiry for elicitation in the P-I plane. This could be a response to prejudice and may increase the expected ego prejudice correction. The rate of increase begins to decline after 3 s and subsequently decreases at 4 s, possibly due to the ego acceleration that occurred at the 3-s mark. 
Later, at 8 s, the pride of the ego increases. This could be due to its lane change within 3 seconds, as shown in Fig.~\ref*{fig:631 scene}. Because of the lane change, the ego action becomes directive with expression and exudes pride from 9 s to 11 s. 
It can be observed that there are fluctuations in pride and depth of inquiry over a brief period, suggesting a deliberate strategy at play. Meanwhile, they are temporally related. The ego inquiry and prejudice have similar changing processes. 
The P-I plane converges to inattention while the P-P plane converges to alignment. 

This case demonstrates that the proposed P-I and P-P planes can clearly describe reciprocal engagement and mutual engagement in communication through the lens of pride and prejudice. 

%% file: sections/5-5_stats_analys.tex
\subsection{Statistical Analysis}
\label{sec4_E}

\textcolor{black}{This study focuses on 42 carefully selected lane-change scenarios that consider duration, trajectory completeness, and interaction quality, and examines the key interactions between the ego vehicle and the designated object vehicle.} Among them, 40 scenarios are categorized into two groups for the statistical analysis: discretionary lane changes (16 instances) and mandatory lane changes (24 instances). These categorized scenarios are adopted to calibrate the personalized driver model with both MIRL and the proposed C-MIRL. 
Additionally, the presumed prior belief distribution in the mandatory lane changes is chosen as a uniform distribution, while that of discretionary lane changes is a bimodal distribution. 

\begin{figure}
    \centering
    \includegraphics[width=0.8\linewidth]{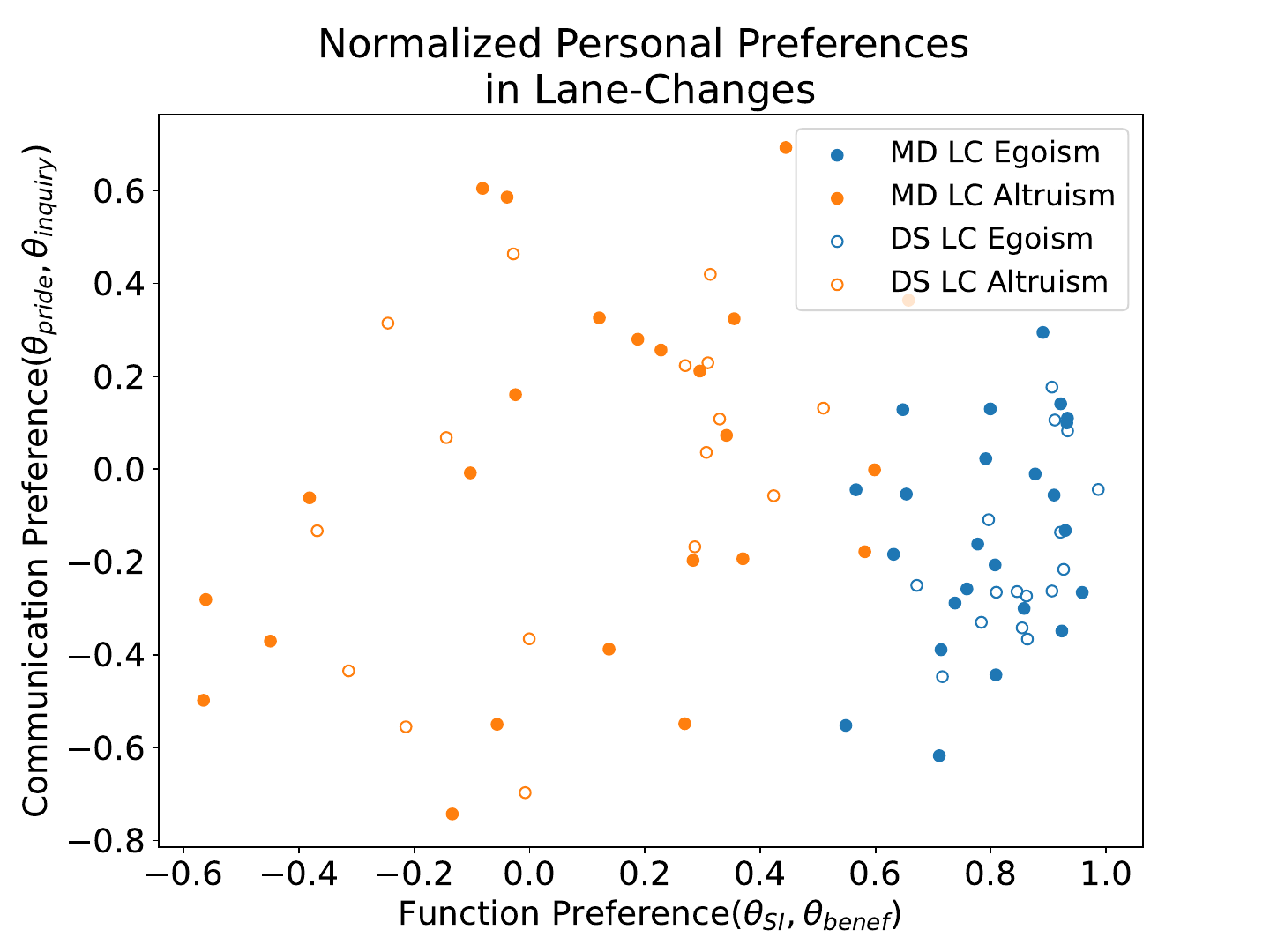}
    \caption[short]{The calibrated results of the personal communication preference weights $\boldsymbol{\theta_{sc}}$ for mandatory lane changes (MD LC) and discretionary lane changes (DS LC). 
    The Egoism utility parameters encompass the weight for self-interest $\theta_{SI}$ (functionality) and pride $\theta_{pride}$ (communication), while the altruism utility parameters refer to the weights of beneficence $\theta_{benef}$ (functionality) and inquiry $\theta_{inquiry}$ (communication).
    }
    \label{fig:theta_sc}
\end{figure}

The calibrated $\boldsymbol{\theta_{sc}}$ of the discretionary and mandatory lane-changing scenarios are shown in Fig. \ref*{fig:theta_sc}. 

Three potential conclusions can be drawn as follows. 
\begin{enumerate}


\item  The mean of the $\theta_{pride}$ of the mandatory results,  -0.141, is bigger than that of the discretionary scenarios, -0.184. Meanwhile, the variance of the $\theta_{pride}$ in the mandatory scenarios, 0.056, is also bigger than that of the discretionary cases, 0.031. Especially, there are only two cases where the $\theta_{pride}$ is positive for the ego driver in discretionary lane changes. 
Our findings indicate that individuals generally do not opt for clear expression in implicit communication when making lane changes.  

\item The mean and variance of $\theta_{inquiry}$ are both larger than those of $\theta_{pride}$. This indicates that inquiry preference is more important while also being more variable among people. Meanwhile, the mean values of $\theta_{inquiry}$ for mandatory and discretionary lane changes are near zero, -0.006 and -0.026, respectively. These results suggest that there is no significant distinction between mandatory and discretionary lane-change cases in terms of inquiry preference. 

\end{enumerate}

%% file: sections/5-6_comparison.tex
\subsection{Comparison Experiments}
\label{sec4_G}

\subsubsection{Compare Active Learning Utilities}

The optimization target is crucial for the active learning process. Previous research established its optimization goal as maximizing the worst-case minimum of the KL divergence between the probabilities of object actions~\cite{yu2023-activeinverselearning}. In the lane-changing context, this corresponds to the difference between the object vehicle action probabilities with adjacent histograms of $\theta_O^{\text{THWF}}$. This could result in sample impoverishment. Conversely, a maximum KL divergence inquiry model is also examined for comparison. 

We adopt these two methods to substitute for inquiry reward for comparison. 

We show the statistical results in Table \ref*{table:md ablation} and Table \ref*{table:dis ablation}. The proposed model demonstrates advantages in mandatory lane changes, and the maxKL inquiry obtains better results than the minKL inquiry. The training results from C-MIRL improved by 19.908\% in comparison to MIRL, while the testing results improved by 14.694\%. The mean of the C-MIRL training results error is 1.400m, while that of the testing results is 1.463m, which indicates the good generalization ability of the model. 

Although maxKL inquiry has a slightly better testing result in discretionary lane changes
, it induces a collision at 7.0s for vehicle 500. This could be explained by the fact that the maxKL definition does not consider the low possibility of extreme situations. Therefore, the maxKL inquiry reward is bigger than that of the proposed mutual information model, which may induce more aggressive inquiry behavior. 



In conclusion, the proposed C-MIRL demonstrates a clear advantage in terms of human driver calibration while maintaining safety, particularly within mandatory lane-changing situations where communication is more committed and engaged.

\subsubsection{Prior Distributions Comparison} 
\label{ssub:prior-distributions-comparison}
\textcolor{black}{
We compare four candidate priors for the ego and object beliefs: beta, normal, bimodal, and uniform. 
}

\textcolor{black}{
The appropriate prior depends on the interaction type. A uniform prior is the natural choice when the ego has no reason to prefer any object preference value. A normal prior is reasonable when most rewards cluster near one typical value. Beta and bimodal priors are more suitable when reward tendencies polarize toward low and high values, which is common in planning-oriented interactions \cite{ramachandran2007bayesian}. 
}

In the proposed information-theoretic communication framework, 
different candidates of the pride and prejudice prior distributions are calibrated, as shown in Tab.~\ref*{table:md prior} and Tab.~\ref*{table:dis prior}. 
Tab.~\ref*{table:md prior} shows that the uniform prior is the most accurate for mandatory lane changes. Tab.~\ref*{table:dis prior} shows that beta and bimodal priors are better for discretionary lane changes. This comparison supports the choice of priors used in the main experiments instead of treating them as arbitrary assumptions. 

\begin{table}
    \footnotesize
    \caption{The comparison study of prior distributions with mandatory lane-changing scenarios. $H_{md}$ is the cross-entropy loss of the mandatory lane-change scenarios. }
    \centering
    \begin{tabular}{lllllll}
        \hline
        Prior           & $FDE_{md-tr}$ & $FDE_{md-te}$ & $H_{md}$  \\
        \hline
        uniform         &    \textbf{1.400}            &   \textbf{1.463}              &     \textbf{2.570}               \\
        normal          &      1.554          &      1.633           &        2.682             \\
        bimodal         &      1.447          &       1.5          &        2.621            \\
        beta    &      1.452          &       1.505          &         2.622           \\
        \hline
    \end{tabular}   
\label{table:md prior}
\vspace{-3mm}
    \end{table}

\begin{table}
    \footnotesize
    \caption{The comparison study of prior distributions with discretionary lane-changing scenarios. $H_{dis}$ is the cross-entropy loss of the discretionary lane-change scenarios. }
    \centering
    \begin{tabular}{lllllll}
        \hline
        Prior           & $FDE_{dis-tr}$ & $FDE_{dis-te}$ & $H_{dis}$  \\
        \hline
        uniform         &    1.04            &   \textbf{1.041}             &     2.39                \\
        normal          &      1.13          &      1.116           &        2.469             \\
        bimodal         &      \textbf{1.005}          &       1.068          &        2.359            \\
        beta    &      1.008          &       1.068          &    \textbf{2.358}           \\
        \hline
    \end{tabular}
    \vspace{-5mm}
\label{table:dis prior}
    \end{table}


%% file: sections/5-7_ablation_study.tex
\subsubsection{Ablation Experiments}
\label{sec4_F}

To analyze the importance of the social-compatibility utilities and affordances, an ablation study for these four features is conducted in Table \ref*{table:md ablation} and Table \ref*{table:dis ablation}. The contributions of the four components and two affordances are examined by dropping one of them at a time. 

\textcolor{black}{Here, MIRL is the non-communicative baseline, C-MIRL is the proposed epistemic-uncertainty-aware communication model, and UA-MIRL (Uncertainty Aware MIRL) is a combined uncertainty-aware model that further incorporates the aleatoric-uncertainty-aware mechanism of our previous work \cite{11048679}. Compared with MIRL, C-MIRL consistently improves the calibration results, which verifies the effectiveness of the proposed communication modeling under epistemic uncertainty. UA-MIRL further improves over C-MIRL in both mandatory and discretionary lane changes, indicating that aleatoric- and epistemic-uncertainty handling contribute complementary gains.} 

It can also be claimed that all the communication features and affordances contribute to improving human likeness. 
Except for discretionary lane changes, the model without an expression affordance shows lower FDE results. Although we model the implicit communication of discretionary lane changes with mandatory preference information, the baseline of MIRL with the mandatory reward beats that without it. Further, the C-MIRL result is significantly improved. 

In analyzing the two communication rewards, it is evident that inquiry rewards play a pivotal role in both scenarios. The model shows the highest FDEs and cross-entropy when the inquiry reward is absent. 
Conversely, self-respect becomes more prominent in mandatory lane changes. 

The listening affordance plays a more significant role than the expression affordance in both scenarios of mandatory and discretionary lane changes. The discretionary lane change model may benefit from the omission of the affordance consideration. From this phenomenon, we can deduce that drivers who engage in discretionary lane changes are more attuned to the tangible costs of pride exudation rather than the presumed expression affordance. 

\begin{table}
    \centering
    \begin{tabular}{lllllll}
        \toprule
        Model           & $FDE_{md-tr}$ & $FDE_{md-te}$ & $H_{md}$  \\
        \midrule
        MIRL            &   1.748      &       1.715    &       2.945              \\
        C-MIRL         &    1.400            &   1.463              &     2.570                \\
          \textcolor{black}{UA-MIRL  }       &    \textbf{1.216}            &   \textbf{1.276}              &     \textbf{2.458}                \\
        W/o Pride         &      1.549          &       1.589          &        2.630            \\
        W/o Inquiry    &      1.582      &      1.574      &       2.689  \\
        minKL Inquiry         &      1.566          &       1.525          &        2.678            \\
        maxKL Inquiry         &      1.506          &       1.496          &        2.599            \\
    
        W/o Expression Affordance          &    1.420            &      1.496           &         2.573            \\
        W/o Listening Affordance         &       1.521         &       1.526          &         2.651           \\
        \bottomrule
    \end{tabular}
    \caption{The ablation study of social compatibility utilities with mandatory lane-changing scenarios. $H_{md}$ is the cross-entropy loss of the mandatory lane-change scenarios. }
\label{table:md ablation}
    \end{table}

    \begin{table}
        \centering
        \begin{tabular}{lllllll}
            \toprule
            Model           & $FDE_{dis-tr}$ & $FDE_{dis-te}$ & $H_{dis}$  \\
            \midrule
            MIRL dis            &   1.137     &   1.146        &      2.499               \\
            MIRL md            &   1.136     &   1.132        &      2.549               \\
            C-MIRL         &    1.005            &      1.068           &     2.359                \\
            \textcolor{black}{UA-MIRL }        &    \textbf{0.936}            &   1.041              &     \textbf{2.246}  \\
            W/o Pride          &    1.024            &      1.057           &     2.389              \\
            W/o Inquiry    &      1.085       &     1.123       &       2.459     \\
            minKL Inquiry          &    1.068            &      1.076           &     2.421              \\
            maxKL Inquiry          &    1.006            &      \textbf{1.039}           &     2.365              \\
            W/o Expression Affordance          &    \textbf{0.991}            &      1.042           &         \textbf{2.335}           \\
            W/o Listening Affordance         &      1.020          &      1.057           &      2.395              \\
            \bottomrule
        \end{tabular}
        \caption{The ablation study of social compatibility utilities with discretionary lane-changing scenarios. $H_{dis}$ is the cross-entropy loss of the discretionary lane-change scenarios. The MIRL dis model is a discretionary lane-change model that does not include a mandatory lane-change reward $R_{md}$. This model is calibrated using MIRL. In contrast, the MIRL md model incorporates a mandatory lane-change reward. }
        \label{table:dis ablation}
    \end{table}

%% file: sections/5-8-HIL.tex
\subsection{Subjective Validity Test with DIL}

To validate that the proposed pride and prejudice are aligned with the actual implicit communication results, subjective evaluations with a pair of Driver-In-the-Loop (DIL) driving simulators are conducted. In the simulators, the participants interact with given merging scenes, and their subjective assessments from both sides are gathered with questionnaires. The consistency between the questionnaire results and the calibrated results of the proposed C-MIRL is tested. 

\subsubsection{Simulator}

\begin{figure}
    \centering
    \includegraphics[width=\linewidth, trim=80 20 80 20, clip]{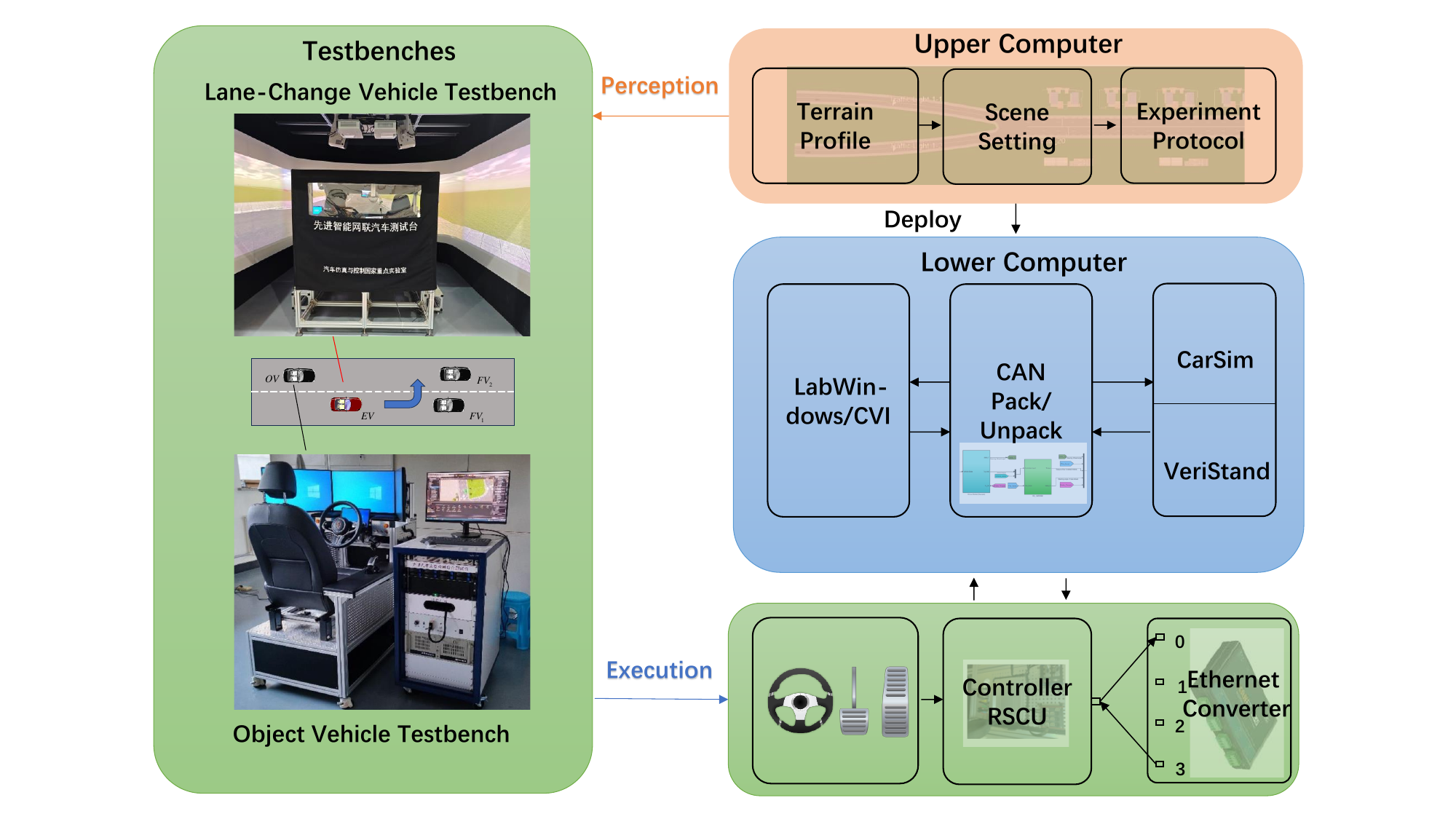}
    \caption[short]{The simulator setup 
    }
    \label{fig:DIL}
\end{figure}

Driving simulators (DS) are crucial platforms for rigorously testing transportation systems and vehicles, providing a secure and controllable experimental environment \cite{zhang2025-drivingsimulatorvalidation}. Fig.\ref*{fig:DIL} provides an overview of the driving simulator's setup.

This platform comprises three essential elements: the testbenches, the upper computer, and the lower computer. 

Two DIL simulator testbenches are employed for the lane-change ego vehicle and object vehicle in a lane-change scene, respectively. The lane-change vehicle testbench adopts a medium-sized fixed-base semi-driving cabin as a simulated vehicle. It is surrounded by a set of immersive projection screens for nearly 260$^{\circ}$  horizontal field of view. 
Four additional screens are adopted for rear views and the dashboard. This configuration is implemented to enhance the participation of drivers in experiments and their sense of presence \cite{almallah2021-drivingsimulationsickness}. 
The object vehicle testbench adopts a triple monitor setup. The advanced features of the lane change simulator stem from the intricate maneuvers necessary for the lane change vehicle. Additionally, subjective evaluations can play a critical role in enhancing the autonomous driving system through a mutual evaluation process. 
The testbenches capture the control output of the drivers with steering wheel, gas, and brake pedals, and broadcast the signals with an Ethernet converter. To execute the steering wheel and brake pedal force feedback from the lower computer, the RSCU controller is adopted. 

The lower computer integrates CarSim and VeriStand with LabWindows/CVI through CAN to create a real-time vehicle dynamics simulation and deploy the scene simulation of the upper computer. 

The upper computer sets up the scene with SCANeR studio. The setting involves a pair of two-lane roads merging into a bidirectional four-lane road running east and west. At this intersection, a traffic light regulates the two vehicles to form experimental scenes on the main road. 
Experimental scenes are encoded with distance and velocity conditions to replicate extracted scenes from NGSIM. 
Within this regulated environment, we can meticulously replicate driving scenarios, guaranteeing uniformity and the ability to reproduce results accurately. 


\subsubsection{Subjective Evaluation}

10 drivers are divided into 5 groups to interact at critical scenes and provide subjective evaluations for each experiment. 
Participants were tasked with completing a questionnaire to assess the conduct of the interacting vehicles following each scenario. The questionnaire comprised six subjective inquiries regarding: 1. role in the interaction, 2. intended expression (exudation of pride), 3. received information (elimination of prejudice), 4. road rage, 5. trust, and 6. trustworthiness in the other vehicle, as depicted in Table \ref*{tab:questionnaire}.
The surveys utilized a 10-point Likert scale to measure the participants' perceptions, except for the interaction role, which was evaluated using binary categorization. This design of a mutual evaluation from both sides allows for thorough subjective validity of the proposed implicit communication model of interaction. In this context, we calculate pride and prejudice according to Eq. (\ref{eq:sc pride}), and then extract their maximum values during interactions for subjective validation. 

\begin{table}[ht]
\centering
\caption{Questionnaire to evaluate the merging vehicle's behavior}
\begin{tabular}{l}
\hline
\textbf{Questions} \\ 
\hline

Q1. Which vehicle did you drive, the lane-changing vehicle \\or the vehicle proceeding straight? \\

Q2. To what extent would you agree that you have clearly expressed \\your intent with driving behaviors to the other vehicle? \\ 
Q3. To what extent would you agree that you have a clear judgement \\on the driving style and intent of the other vehicle? \\ 
Q4. To what extent would you agree that 
the object vehicle was \\experiencing road rage? (not yielding or not slowing down)\\
Q5. To what extent would you trust the other vehicle? \\ 
Q6. To what extent would you agree that the object vehicle is\\ trustworthy? (Understanding of the object preferences and collaboration) \\ 
\hline

\end{tabular}
\label{tab:questionnaire}
\end{table}

\subsubsection{Validation Results}

\begin{figure}
    \begin{minipage}[b]{0.45\linewidth}
        \subcaptionbox{\label{fig:css_prd}}{
        \includegraphics[width=\linewidth, trim=170 280 180 290, clip]{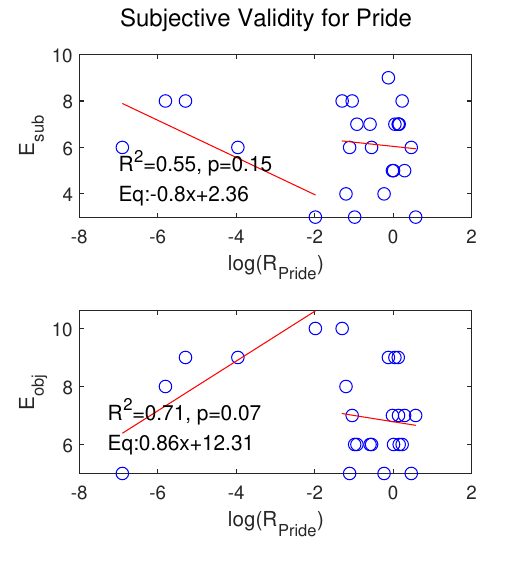}
        }
    \end{minipage}
    \begin{minipage}[b]{0.45\linewidth}
        \subcaptionbox{\label{fig:css_prj}}{
        \includegraphics[width=\linewidth, trim=170 280 180 290, clip]{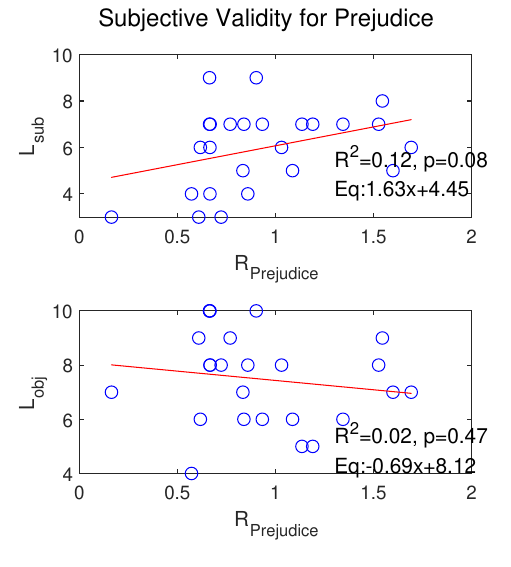}
        }
    \end{minipage}
    \caption[short]{(a) Subjective validity results of pride, grouped with the threshold of -1.5. (b) Subjective validity of prejudice. }
    \label{fig:consistency}
\end{figure}

First, correlation analysis was carried out to examine the relationship between subjective evaluation criteria and the objectively calibrated pride and prejudice. Fig.\ref*{fig:consistency} illustrates the relationship between mean expression clarity and the logarithm of pride values. The plot shows a positive correlation for the logarithm of pride value below -1.5, supported by a regression line with the equation $0.86x+12.31$. The model explains a substantial portion of the variance ($R^2 = 0.71$) and suggests marginal statistical significance (p = 0.07). 
The complex relationship involving pride may stem from the fact that heightened levels of pride can lead to road rage, ultimately impacting the way in which one expresses oneself. This can result in a compromised clarity of expression. 

To further investigate potential interactions between road rage and pride, a multivariate regression analysis with an interaction term and mediation analysis is employed as shown in Table~\ref*{table:mlti var reg}. The group is represented by a boolean value determined by whether the logarithm of pride is greater than -1.5. 
The significant group coefficient ($p = 0.02$) suggests systematic differences between groups, while the interaction hints that pride's impact may vary by group. Meanwhile, higher values of road rage significantly reduce expression clarity ($\beta=-0.406, p=0.003$). The multivariate regression model explains 54.8\% of the variance in expression clarity ($R^2 = 0.548$) and is highly significant overall ($p = 0.002$). 

\begin{table}
    \footnotesize
    \caption{Result of multivariate regression: subjective expression clarity}
    \centering
    \begin{tabular}{lllllll}
        \hline
        Parameter      &$\beta $    & $df$ & $F-value$ & $p-value$  \\
        \hline
        pride         &0.62&    1            &   3.69              &     \textbf{0.07}                \\
        group         &-4.19&      1          &       6.26          &       \textbf{0.02}            \\
        road rage    &-0.41&      1     &     11.40      &      \textbf{0.003}  \\
        pride*group         &-1.12&      1          &       4.14          &        \textbf{0.06}           \\
        \hline
    \end{tabular}
\label{table:mlti var reg}
    \end{table}

To explore the underlying mechanisms driving the observed group differences and the mediation effects of road rage, a separate mediation analysis for each pride group is conducted. 

\begin{table}
    \footnotesize
    \caption{Result of mediation analysis: X: pride, Y: subjective expression clarity, M: road rage}
    \centering
    \begin{tabular}{l|ll|ll}
        \hline
        Parameter      & \multicolumn{2}{c|}{$\log(R_{pride})<-1.5$} & \multicolumn{2}{c}{$\log(R_{pride})\ge-1.5$}\\
        \cline{2-5}
        &$\beta $    & $p-value$ & $\beta$ & $p-value$  \\
        \hline
        direct effect         &1.04&    0.59           &   -0.61             &     0.18              \\
        indirect effect        &-0.17&      0.86          &       0.39         &       0.44            \\
        total effect   &0.87&      \textbf{0.016}     &    -0.22     &      0.89  \\
        \hline
    \end{tabular}
\label{table:mediatn1}
    \end{table}

 Mediation analysis stratified by pride levels revealed a significant total effect of pride on expression clarity only for individuals with low pride ($\log(R_{pride}) < -1.5; \beta=0.87,p=0.016$), though no significant indirect effects were detected. In contrast, the high-pride group showed no significant mediation, despite a strong standalone mediator-outcome relationship ($p<0.001$). This suggests that pride level moderates pride's impact, with mechanisms differing across subgroups. At higher levels of pride, road rage becomes more significant.

On the other hand, the mean listening received information and the prejudice calibrated have a correlation coefficient of $R^2 = 0.12$ and a p value of 0.08 as shown in Fig. \ref*{fig:css_prj}. The bigger variance indicates that the recognition performance may vary widely among listeners. Further, the subjective evaluation can be better fitted for the receiver, which, conversely, surpasses the sender. This indicates the effectiveness of a mutual evaluation. 


%% file: sections/6-discussion.tex
We have developed a driver behavior model that facilitates mutual communication through context-based information preference. This work advances information-theoretic modeling of vehicle implicit communication.

Mutual communication information is quantified with context-based information channels, namely pride and inquiry. Ego inquiry is the expected prejudice over its action probability. Therefore, we adopt inquiry to define epistemic uncertainty. 
Pride and inquiry are adopted to form a Pride-Inquiry (P-I) plane and a Pride-Prejudice (P-P) plane. Considering communication as the prejudice correction and pride exudation process, two communication planes describe the intensity and tendency of one-sided engagement and mutual engagement. 

Verified on the NGSIM dataset with C-MIRL, the training results for the mandatory-lane-change driver behavior model are improved by 19.908\%, while the testing results are improved by 14.694\%. We find that people seldom prefer to convey clear messages through implicit communication during lane changes, especially discretionary lane changes. 
Individuals with higher levels of pride preference may be more
likely to participate with commitment in engaged situations 
while also showing more variance in mandatory lane changes.  On the other hand, inquiry preference is important to both
kinds of lane changes, while also being more variable among
people. Given that inquiry serves as a method to measure epistemic uncertainty, it can be inferred that reluctance toward inquiry is more closely tied to personal characteristics than the expression of pride or aggression \cite{allport1979nature}. 

Further, the correlation between objectively calibrated results and subjective evaluations from the DIL experiment questionnaires is analyzed. It confirms significant positive correlations between subjective expression and the logarithm of pride at a lower level of pride. At a higher level of pride, road rage becomes significant and compromises expression clarity. Additionally, significant positive correlations are found between listening and the prejudice corrected.
The higher variance in subjective listening clarity, linked to calibrated prejudice in regression analysis, suggests wide variability in recognition performance, akin to speech recognition \cite{samardzic2023-utilizationpsychometricfunctions}. 
Besides, in both directions of communication, the mean of subjective evaluation can be better fitted for the receiver, which, conversely, surpasses the sender. This indicates the effectiveness of a mutual evaluation.

The information-theoretic framework of implicit mutual communication among vehicles is multifaceted, incorporating both contextualized transmission channels and opportunities for decontextualized analysis of information exchange processes. This not only opens up opportunities for vehicle interactions at intersections and roundabouts but also provides a basis for other human-automation interactions to fight pride and prejudice with communication, such as shared steering control \cite{hu2024-trustbasedsharedcontrol} and the large language model (LLM) alignment problem \cite{peschl2022-moralaligningai, yao2024-claveadaptiveframework}. 

The proposed model can help autonomous vehicles fight discrimination and bullying through communication while navigating unknown interactive environments. It quantifies epistemic uncertainty and evaluates communication intensity and tendency. %
Future work could improve computational efficiency for online decision making. Learning-based world models could be adopted to replace the simulation environment.

%% file: appendix/1-inquiry-as-eu.tex
Inquiry is the expected prejudice over possible object actions.
\begin{equation}
    \begin{split}
        I(\zeta_O;\theta_O)=&I(\theta_O;\zeta_O)\\
        =&
                \mathbb{E}_{\zeta_O}
                \left(
                D_{K L}
                    \left[P_{L_1}\left(\theta_O \mid \mathbf{x}^0, \tilde{\zeta}^j\right)
                    \| P_{0}(\theta_O)
                    \right]\right)
        \end{split}
    \label{eq:in as clar}
\end{equation}

Therefore, the ego inquiry is adopted as the measurement of the epistemic uncertainty of the ego and vice versa. 

While higher inquiry gains more information, it also introduces more epistemic uncertainty. This could be mitigated with stereotyping. 
Stereotyping can be defined as the entropy of the prior beliefs. 
\begin{equation}
    R_{stereo} = H(\theta) = -\sum_{q=1}^{N_O}P_0(\theta)\log P_0(\theta)
    \label{eq:stereo}
\end{equation}
Conversely, the mutual information model of epistemic uncertainty can be rewritten with entropy. 
\begin{equation}
    I(\zeta_O;\theta_O)=H(\theta_O)-H(\theta_O\mid \zeta_O )
    \label{eq:anxiety}
\end{equation}
Therefore, stereotyping can reduce epistemic uncertainty by limiting its upper bound. It is worth noting that this process could also introduce prejudice. 